\documentclass[12pt]{article}
\usepackage{filecontents}
\usepackage{amsmath}
\usepackage{graphicx}
\usepackage{enumerate}
\usepackage{natbib}
\usepackage{url} 
\usepackage{multirow}
\usepackage{algorithm,algpseudocode}
\usepackage{amsthm}
\usepackage{xcolor}
\usepackage{setspace}
\usepackage[hidelinks]{hyperref}
\hypersetup{
    colorlinks,
    linkcolor={red!50!black},
    citecolor={blue!50!black},
    urlcolor={blue!80!black}
}
\newtheorem{definition}{Definition}
\newtheorem{theorem}{Theorem}

\newtheorem{assumption}{Assumption}
\newtheorem{remark}{Remark}
\usepackage{enumitem}

\usepackage{amsmath,amsfonts,bm}

\def\eqref#1{equation~\ref{#1}}

\def\1{\bm{1}}

\DeclareMathAlphabet{\mathsfit}{\encodingdefault}{\sfdefault}{m}{sl}
\SetMathAlphabet{\mathsfit}{bold}{\encodingdefault}{\sfdefault}{bx}{n}

\newcommand{\E}{\mathbb{E}}

\renewcommand{\P}{\mathbb{P}}

\newcommand{\bx}{\boldsymbol{x}}

\algrenewcommand\algorithmicindent{1em}%

\newcommand{\blind}{0}

\newcommand{\indep}{\;\, \rule[0em]{.03em}{.67em} \hspace{-.27em}
	\rule[-.02em]{.7em}{.03em} \hspace{-.27em}
	\rule[0em]{.03em}{.67em}\;\,}

\graphicspath{{../}}
\usepackage{array}

\usepackage{xr}
\makeatletter
\newcommand*{\addFileDependency}[1]{
  \typeout{(#1)}
  \@addtofilelist{#1}
  \IfFileExists{#1}{}{\typeout{No file #1.}}
}
\makeatother

\newcommand*{\myexternaldocument}[1]{
    \externaldocument{#1}
    \addFileDependency{#1.tex}
    \addFileDependency{#1.aux}
}

\myexternaldocument{CFRL_supp}

\renewcommand{\baselinestretch}{0.8}

\begin{document}

\def\spacingset#1{\renewcommand{\baselinestretch}%
{#1}\small\normalsize} \spacingset{1}


\if1\blind
{
  \title{\bf Counterfactually Fair Reinforcement Learning via Sequential Data Preprocessing}
} \fi

\if0\blind
{
  \bigskip
  \bigskip
  \bigskip
    \title{\bf Counterfactually Fair Reinforcement Learning via Sequential Data Preprocessing}
    \author{  Jitao Wang\textsuperscript{1,a}, Chengchun Shi\textsuperscript{2,b}, John D. Piette\textsuperscript{3,c}, \\
      Joshua R. Loftus\textsuperscript{2,d}, Donglin Zeng\textsuperscript{1,e}, Zhenke Wu\textsuperscript{1,f}\\
      \\
        \textsuperscript{1}Department of Biostatistics, University of Michigan, USA\\
        \textsuperscript{2}Department of Statistics, London School of Economics, UK\\
        \textsuperscript{3}Department of Health Behavior and Health Equity, \\School of Public Health, University of Michigan, USA \\
        \textsuperscript{a}\href{mailto:jitwang@umich.edu}{\small jitwang@umich.edu},
        \textsuperscript{b}\href{mailto:c.shi7@lse.ac.uk}{\small c.shi7@lse.ac.uk},
        \textsuperscript{c}\href{mailto:jpiette@umich.edu}{\small jpiette@umich.edu},\\
        \textsuperscript{d}\href{mailto:j.r.loftus@lse.ac.uk}{\small j.r.loftus@lse.ac.uk},
        \textsuperscript{e}\href{mailto:dzeng@umich.edu}{\small dzeng@umich.edu},
        \textsuperscript{f}\href{mailto:zhenkewu@umich.edu}{\small zhenkewu@umich.edu}
    } 
    \medskip
} \fi
\date{}
\maketitle


\bigskip
\begin{abstract}
When applied in healthcare, reinforcement learning (RL) seeks to dynamically match the right interventions to subjects to maximize population benefit. However, the learned policy may disproportionately allocate efficacious actions to one subpopulation, creating or exacerbating disparities in other socioeconomically-disadvantaged subgroups. These biases tend to occur in multi-stage decision making and can be self-perpetuating, which if unaccounted for could cause serious unintended consequences that limit access to care or treatment benefit. Counterfactual fairness (CF) offers a promising statistical tool grounded in causal inference to formulate and study fairness. In this paper, we propose a general framework for fair sequential decision making. We theoretically characterize the optimal CF policy and prove its stationarity, which greatly simplifies the search for optimal CF policies by leveraging existing RL algorithms. The theory also motivates a sequential data preprocessing algorithm to achieve CF decision making under an additive noise assumption.  We prove and then validate our policy learning approach in controlling unfairness and attaining optimal value through simulations. Analysis of a digital health dataset designed to reduce opioid misuse shows that our proposal greatly enhances fair access to counseling.

\end{abstract}

\noindent%
\vfill

\newpage
\spacingset{1.8} 
\section{Introduction}
\label{sec:intro}

With the widespread integration of machine learning (ML)-based decision making systems in various sectors of the economy such as banking, education, financial analysis and healthcare, there is a growing focus on the ethical and societal implications of its deployment \citep{howard2018ugly,chouldechova2018case,mehrabi2021survey}. The susceptibility of ML algorithms to bias is raising concerns about the potential for discrimination, particularly as it affects already socioeconomically-disadvantaged subgroups (e.g., racial/ethnic minorities or women). For example, in healthcare, automated decision making systems may unfairly allocate treatment resources to specific subpopulations to maximize population-wise long-term benefits, while neglecting the needs of patients already at risk for limited access or poor outcomes. To redress these problems, researchers have proposed the imposition of fairness constraints on decision-making in order to achieve various fairness-related objectives \citep{hardt2016equality,chouldechova2017fair,yeom2018discriminative,mehrabi2021survey,black2022algorithmic,corbett2023measure}.

The counterfactual fairness \citep[CF,][]{kusner2017counterfactual} adopted in this paper requires that, all else being equal, the distribution of the decisions for an individual be the same had the individual belong to a different group defined by a sensitive attribute (e.g., gender, race). Unlike other fairness definitions, CF offers a solution based on causal reasoning, which lends itself to statistical techniques to eliminate the influence of sensitive attributes on the outcomes \citep{kusner2017counterfactual,silva2024counterfactual}. \citet{dedeo2014wrong}  further argued that even the most successful algorithms would fail to make fair judgments without causal reasoning. To 
illustrate the differences in major existing fairness concepts, 
consider the following simplified single-stage university admissions example 
\citep{wang2022adjusting}: 
a university wants to develop an ML-based decision support system for undergraduate admission. The system makes admission decisions based on an applicant's information, including entrance exam score, gender, race/ethnicity, access to prior educational opportunities, where the score may be correlated with the rest of the variables and unmeasured factors like aptitude. Below, we discuss three notions of fairness: 
\begin{enumerate}[leftmargin=*]
\item \textit{Demographic parity}: If the university were to pursue ``demographic parity'' with respect to gender, the goal would be to ensure admission of an equal proportion of male and female applicants regardless of entrance exam scores or the student's aptitude. It only seeks to match the proportion of admission between the groups, regardless of actual qualifications, which can be problematic if one group is genuinely different in terms of the distribution of actual qualifications. It is purely a statistical metric without considering causal mechanisms that lead to differences in the observed data. 
\item \textit{Equal opportunity}: If the university were to pursue ``equal opportunity'' \citep{hardt2016equality} with respect to gender, the goal would be to ensure that truly qualified applicants would be admitted with equal probabilities (i.e., true positive rates) across groups defined by gender. Equal opportunity relies on observational measures and checks statistical parity conditional on the true qualification but does not consider why or how group differences arise, i.e., it does not directly model or reason about causal relationships. 
\item \textit{Counterfactual fairness}: CF with respect to gender shifts attention to individual-level and a causal interpretation of fairness. It ensures that the chance of being admitted does not change had their gender been hypothetically switched while keeping all else about the individual's situation the same. Instead of focusing on matching group-level metrics, CF involves building or assuming a causal model that explains how sensitive attributes influence the individual's observed features. 
\end{enumerate}
This example highlights two key distinctions of CF compared to other fairness definitions: 1) CF focuses on individual-level fairness rather than group-level, and 2) it seeks to remove direct and indirect effects of sensitive attributes on decisions, ensuring fairness from a causal, rather than purely statistical, perspective, by considering fairness in slightly different hypothetical worlds.


Significant progress has been made in achieving CF decision making in single-stage scenarios. However, focusing solely on a single static decision can lead to suboptimal outcomes if it fails to consider the dynamic nature of sequential decision-making processes 
\citep{liu2018delayed,creager2020causal,d2020fairness}. Reinforcement learning (RL) excels at sequential decision-making applications including fintech \citep{malibari2023systematic}, traffic light control \citep{wei2018intellilight}, and healthcare \citep{li2022reinforcement,buccinca2024towards}, by learning policies that maximize future rewards. Despite its success, integrating CF principles within RL remains an under-explored area \citep{reuel2024fairness}. The challenges of applying CF in RL are two-fold. First, unlike static scenarios, 
sensitive attributes may not only directly affect the current state, but also indirectly influence the current state through its impact on previous states and decisions. Disentangling these direct and indirect effects to ensure CF is a significant challenge. Second, most existing RL algorithms operate under a Markov decision process (MDP) model assumption, under which the optimal policy achieves desirable properties such as Markovian properties and stationarity \citep{puterman2014markov}. These properties substantially reduce the search space and simplify the algorithm. It is 
unclear whether 
these properties still hold when incorporating CF constraints. 

Our work is motivated by the ``PowerED'' study \citep{piette2023automatically}.  The original study was a randomized controlled trial of patients who were at risk for opioid-related harms (e.g., overdose or addiction) in which investigators evaluated an RL-supported, 12-week digital health intervention designed to prevent those negative outcomes through behavioral counseling of opioid users while conserving scarce counselor time. Specifically, the intervention arm used an RL-based automated decision making system to dynamically assign weekly personalized treatment options based on each patient's self-reported pain score and opioid use behaviors in the prior week, with the goal of reducing self-reported opioid misuse behaviors. Treatment options included i) brief motivational interactive voice response (IVR) call (less than 5 minutes), ii) a longer recorded call (5 to 10 minutes), or iii) a live call with counselor (20 minutes). Comparison-group participants received 12 weeks of standard motivational enhancement via weekly calls with a trained counselor. In this paper, we focus on the RL-supported arm of the trial where an RL agent seeks to intelligently allocate a limited supply of counselors' time (option iii). 

There are two ways unfairness could be introduced in an application like the PowerED study if fairness-unaware RL algorithms are used (as they were in this trial). First, by directly including sensitive attributes (e.g., ethnicity and gender) into the set of state variables within the RL model, the learned policy could make biased decisions for those minority groups characterized by the sensitive attributes. For example, patients with similar pain levels but different ethnicities might receive different treatment options if the agent uses ethnicity as a decision factor. 
Second, even when sensitive attributes are excluded from the state variables to make decisions, the effect of the sensitive attribute upon the state variables may still indirectly influence the agent's decisions. For instance, research indicates that Hispanic Americans, despite experiencing higher pain sensitivity, often report fewer pain conditions due to cultural factors \citep{hollingshead2016pain}. This under-reporting may mislead the agent to think that Hispanics are experiencing less pain compared to other subgroups and to assign fewer human counseling (option iii above; more efficacious) to Hispanic patients, creating unfairness. In this paper, we demonstrate that CF provides a promising statistical framework to study fair sequential decisions while controlling for both direct and indirect influences of the sensitive attributes.

\paragraph{Contributions} Our work makes four main contributions: (i) Motivated by 
a real-world digital-health application, the PowerED study, we propose a novel and generalized CF definition that is suitable for the more challenging but ubiquitous dynamic setting in which RL makes sequential decisions; 
(ii) We theoretically characterize the class of CF policies and prove the stationarity of the optimal CF policy which greatly simplifies the search and evaluation for optimal CF policies by leveraging existing RL algorithms. (iii) Motivated by the theory, we propose a sequential data preprocessing algorithm designed for optimal CF policy learning under an additive noise assumption. We establish theoretical guarantees that our approach asymptotically controls the level of unfairness and attaining optimal value; (iv) We demonstrate the efficacy of our proposed algorithm in controlling unfairness and attaining optimal value through numerical studies and real data analysis.


\paragraph{Related Work on Fair ML} A variety of fairness criteria in ML and their characterizations in the literature have been recently reviewed \citep{kleinberg2018algorithmic,mehrabi2021survey,barocas2023fairness,yang2024survey,caton2024fairness}. Broadly speaking, ML methods addressing fairness goals can be classified into three categories. 
\textit{Preprocessing} approaches remove potential bias from the data before training. For example, \citet{kusner2017counterfactual,nabi2018fair,chiappa2019causal,salimi2019interventional,zuo2022counterfactual,chen2023learning} used the causal framework of directed acyclic graph (DAG) to remove unfairness from the training data. Others proposed to preprocess the training data using relabelling and perturbation techniques to balance between underprivileged and privileged instances \citep{kamiran2012data,jiang2020identifying,wang2019repairing}. \textit{In-processing} methods incorporate fairness metrics directly into the model's training process, for example by regularization or constrained optimization \citep{berk2017convex,aghaei2019learning,di2020counterfactual,viviano2024fair} or with adversarial approaches \citep{edwards2015censoring,beutel2017data,celis2019improved}.
\textit{Post-processing} approaches adjust the model's predictions after training to ensure fairer outcomes for different groups \citep{pleiss2017fairness,hebert2018multicalibration,kim2019multiaccuracy,wang2022adjusting}. Several recent works have explored various approaches to achieve CF in single-stage scenarios. \citet{chen2023learning} proposed an algorithm that removes sensitive information from the training data. \citet{wang2022adjusting} developed a post-processing procedure to make unfair ML models fair. \citet{kusner2017counterfactual} and \citet{zuo2022counterfactual} considered building ML models that rely only on non-sensitive attributes. \citet{di2020counterfactual} proposed to use regularization by incorporating fairness penalty into the loss function.




\paragraph{Paper organization} The remainder of this paper is organized as follows. Section \ref{sec:pre} reviews structural causal models, counterfactual inference, and the contextual Markov decision process (CMDP) model. In Section \ref{sec:cfrl}, we extend single-stage CF to multi-stage decision making under the framework of CMDP. 
In Section~\ref{sec:cf_class}, we characterize the form of (optimal) CF policies 
when the counterfactuals are known. In Section~\ref{sec:learning}, we propose a sequential data preprocessing algorithm for estimating the counterfactuals. The preprocessed data serve as inputs to any offline RL algorithm for CF policy learning. We then theoretically establish value optimality and asymptotic fairness control of the learned policy in Section \ref{sec:theory_policy}, which are empirically supported by synthetic and semi-synthetic experiments  in Section~\ref{sec:num_exp}. We apply the algorithm to a real-world interventional digital health dataset Section~\ref{sec:real_data}. The paper concludes with a brief discussion on limitations and future directions.

\section{Preliminaries} \label{sec:pre}
To bring the discussion of fairness into the framework of causal inference, we first give a brief introduction to the definitions of structural causal model (SCM) and counterfactuals. Then we introduce the the framework of CMDP in which we define the CF using the language of SCM (Section \ref{sec:cfrl}). 

\subsection{Structural causal model and counterfactuals}
\label{ssec:scm}
Following \citet{pearl2000models}, SCM provides a mathematical framework for modeling causal relationships between variables. An SCM $M=(\mathcal{U},\mathcal{V},F)$ consists of a set of observable endogenous variables $V$, a set of unobserved exogenous variables $U$ and a set of functions $F$. These functions assign value to each endogenous variable $V$ given its parents $pa(V) \in \mathcal{V}$ and unobserved direct causes $U$: $V = f_{V}(pa(V),U)$. $U$ are required to be jointly independent. The structure of $M$ can be depicted by a causal graph, usually in the form of directed acyclic graph (DAG), where $(pa(V), U)$ forms the parent set of $V$.

\begin{figure}[ht!]
    \centering
    \includegraphics[width=0.65\linewidth]{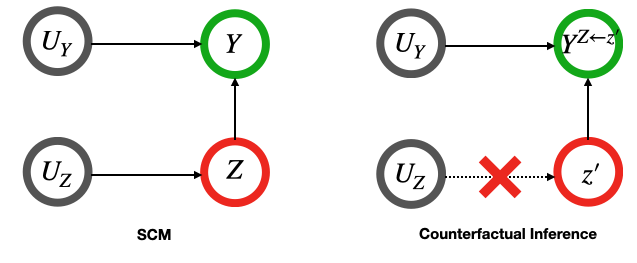}
    \caption{A simple example of SCM and counterfactual inference.}
    \label{fig:scm_ci}
\end{figure}

In the SCM framework, we define counterfactual inference. Assume $Y \in V$ is the variable of interest. Let $Z$ and $U_Y$ be the parents of $Y$ where $Z \in V$ and $U_Y \in U$; let $U_Z$ be the parent of $Z$ and $U_Z\in U$. By the definition of SCM, $Y$ is fully determined by $Z$ and $U_Y$ through function $f_Y$, i.e., $Y=f_Y(Z,U_Y)$. Given a realization of $U_Y=u_Y$, suppose we have observed the values of $Z$ and $Y$, denoted as $z$ and $y$. Consider a typical counterfactual inference statement: what would be the value of $Y$ had $Z$ taken value $z'$ instead of $z$ given we have observed $z$ and $y$?
This counterfactual query can be realized through Pearl’s do-operator, $do(Z = z')$, which generates an interventional distribution by removing the edges leading into $Z$ in the corresponding DAG and set $Z$ to the value $z'$. Following the notation in the seminal CF paper \citep{kusner2017counterfactual}, we denote it by $Y^{Z\leftarrow z'}(U_Y)$, where ``$Z\leftarrow z'$'' represents the $do(Z = z')$ operation and $Y^{Z\leftarrow z'}(U_Y)$ is a deterministic function of $Z$ and $U_Y$. Figure~\ref{fig:scm_ci} depicts the difference between SCM and counterfactual inference.
It is important to note that while $U_Y$ is not directly observable, we can infer its value from the observed values $z$ and $y$. This inference of $U_Y$ is essential for computing counterfactuals.

We adhere to a general three-step procedure for counterfactual inference to compute this quantity \citep[please refer to][ for more details]{pearl_causal_2016}: 1. \textit{Abduction}: update $P(U_Y)$ by the observed quantities $Y=y, Z=z$, obtaining $P(U_Y|Y=y, Z=z)$. 2. \textit{Action}: remove the structure equations for $Z$ and replace them with the appropriate value, i.e. $Z=z^\prime$. 3. \textit{Prediction}: use the modified structural model and updated $P(U_Y|Y=y, Z=z)$ to compute $Y^{Z\leftarrow z'}(U_Y)$. We will detail this inference procedure for our specific settings in Section~\ref{sec:contextualbandit} and \ref{ssec:cf_cmdp}. 

\subsection{Contextual Markov Decision Process}
\label{sec:cmdp}
Contextual Markov Decision Process \citep[CMDP,][]{hallak2015contextual} is an augmented MDP that incorporates additional contextual information during decision making. 
\begin{definition}[CMDP]
	Contextual Markov decision process is a tuple $(\mathcal{C},\mathcal{S},\mathcal{A},\mathcal{M}(c))$ where $\mathcal{C}$ is called the context space, $\mathcal{S}$ and $\mathcal{A}$ are the state and action space, and $\mathcal{M}$ is function mapping any context $c\in\mathcal{C}$ to an MDP $\mathcal{M}(c) = (P^c,R^c)$.
\end{definition}

\begin{figure}[ht!]
    \centerline{\includegraphics[width=0.75\textwidth]{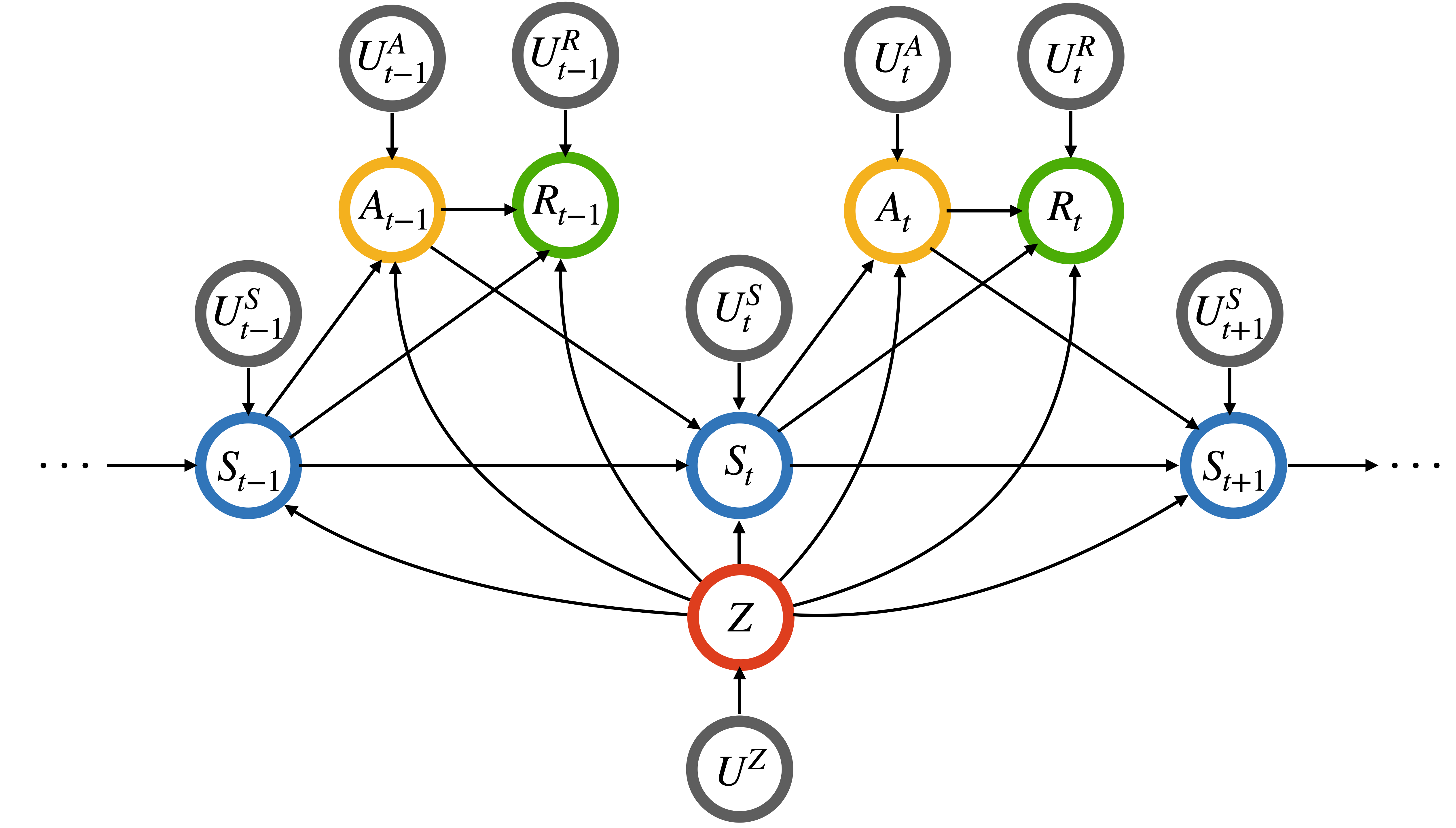}}
    \caption{Causal DAG of CMDP.} 
    \label{fig:CMDP}
\end{figure}

We assume the observational data follows a CMDP where the sensitive attributes serve as the contextual information, as illustrated in Figure~\ref{fig:CMDP}. Consider the collected observed tuples $\{(z_{i},s_{it},a_{it},r_{it}):t=1,\dots,T_i, i=1,\dots,N\}$, where $N$ denotes the number of subjects and $T_i$  denotes the length of horizons for subject $i$. For simplicity, we fix $T_i = T$ for all $i$ in the rest of the paper.
$Z_i$ denotes time-invariant sensitive attributes for subject $i$, which takes value from $\mathcal{Z} = \{z^{(1)},\dots,z^{(K)}\}$. In this paper, we consider the sensitive attributes to be categorical, which is a reasonable assumption for commonly studied attributes such as race and gender. Additionally, for ease of presentation, we focus on a single sensitive attribute in this paper, but the extension to multiple attributes is straightforward. For simplicity, we omit index $i$ for individual. Let $S_{t}, A_{t}, R_{t}$ represent the state variables, actions and received reward at time $t$. $U^S_{t}, U^R_{t}$ are the exogenous variables for $S_{t}, R_{t}$. We use ${\pi} = \{\pi_t\}_{t\geq1}$ to denote a policy, which consists of a sequence of decision rules. At time $t$, the environment arrives at a state $S_{t}$ and the agent takes an action $A_{t}$ based on a behavior decision rule $\pi_t^{(b)}$ which is specific to the mechanism of how the observed data at hand are collected and often differ from an alternative and potentially better decision rule $\pi_t$. The environment then transitions to a new state $X_{t+1}$ and gives a reward $R_{t}$ according to transition kernel $P_t(S_{t+1},R_{t} | S_{t},A_{t},Z)$. Here the transition kernel $P_t$ can vary by time in general as indicated by the dependence on the index $t$; $\pi_t$ can generally depend on entire history $H_t=\{Z,\bar{S}_t, \bar{A}_{t-1}, \bar{R}_{t-1}\}$ where the notation ``$\bar{W}_t$'' generically represents the sequence of variable $W$ from time $1$ up to and including time $t$.

We introduce two assumptions as implied by the DAG structure in the CMDP framework and remark on their relevance to our theoretical analyses. 
\begin{assumption}[No unmeasured confounders]\label{asm:unmeasured_confounders}
    For each $t<T$, conditional on $H_t$ blocks all backdoor paths from $A_t$ to $S_{t+1}$ and from $A_t$ to $R_t$.
\end{assumption}

\begin{assumption}[Markov property]\label{asm:markov}
    For any $t<T$, $S_{t+1}, R_t \indep \{S_j,R_{j-1},A_j\}_{j\leq t-1} \mid S_{t}, A_t, Z$
\end{assumption}

\begin{remark} Assumption~\ref{asm:unmeasured_confounders} ensures that, conditional on the history $H_t$, there exists no unmeasured confounders between the action $A_t$ and the subsequent state-reward pair $(S_{t+1}, R_t)$. It is automatically satisfied in our PowerED study where the behavior policy is RL-based and relies solely on patient's  observed information. 
This condition is crucial as it enables consistent estimation of transition and reward functions from observational data. Notably, when this assumption is violated, it may compromise the Markov property, resulting in a confounded partially observable MDP \citep{lu2022pessimism,shi2022minimax,bennett2024proximal,hongmodel}. Assumption~\ref{asm:markov} implies that the next state $S_{t+1}$ and reward $R_t$ following action $A_t$ are conditionally independent of the entire history given the current state $X_t$, action $A_t$, and sensitive attribute $Z$, making the transition and reward functions only dependent on $S_{t}, A_t, Z$ rather than the entire history. Consequently, this Markov property enables more efficient policy learning as decisions can be made based on the current state.
\end{remark}


\section{Counterfactual Fairness in RL}
\label{sec:cfrl}
Before extending single-stage CF \citep{kusner2017counterfactual} to complex CMDP settings, we first consider a simpler contextual bandit setting, which is a simplification of CMDP with only one time step; it helps establish the meanings of notation and the heuristics of why preprocessing strategies can help achieve CF.

\subsection{Counterfactual fairness under contextual bandit}
\label{sec:contextualbandit}

\begin{figure}[ht!]
	\centering
	\includegraphics[width=0.55\textwidth]{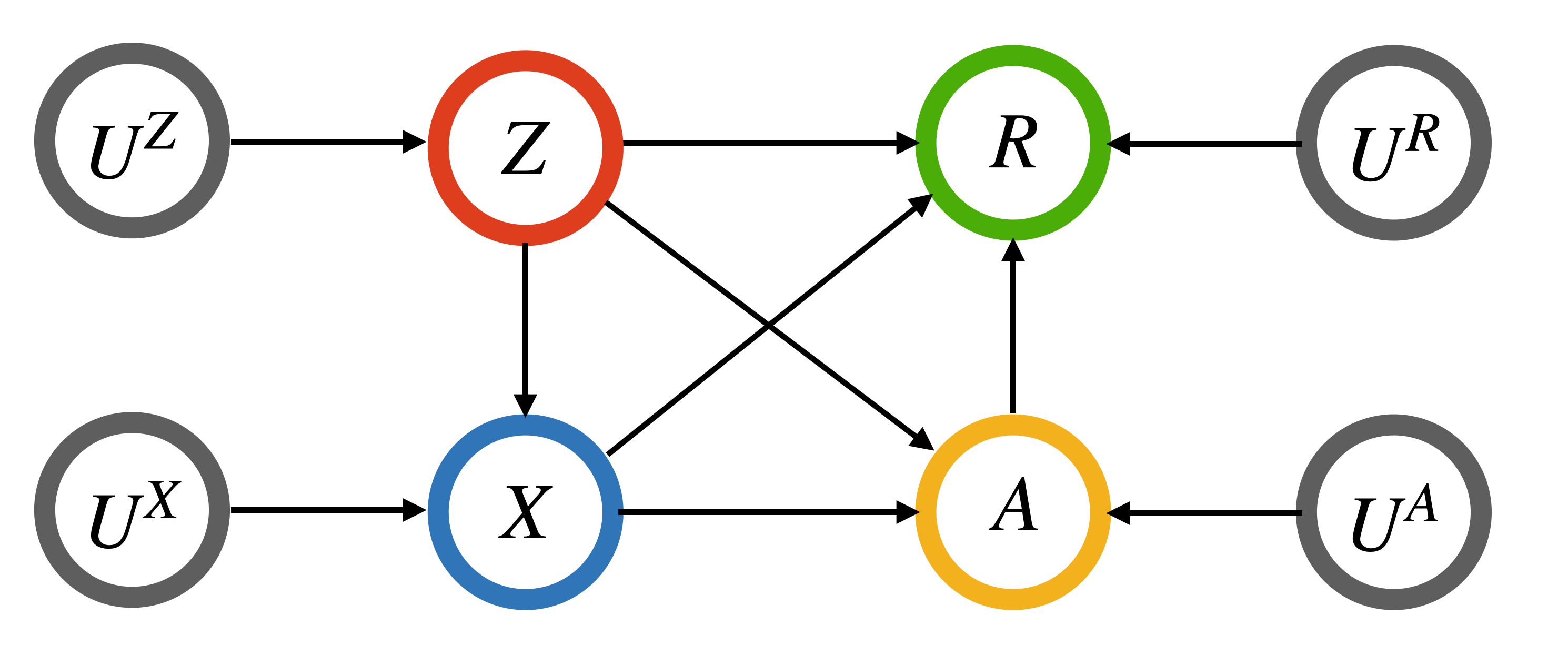}
	\caption{Causal DAG of contextual bandit (one-stage).}\label{fig:CB}
\end{figure}

Let $(Z,X,A,R)$ be the variables in a contextual bandit setting, as illustrated in Figure~\ref{fig:CB}. Here, $Z$ denotes the sensitive attribute, $X$ represents the non-sensitive context (of dimension one for ease of presentation here), $A$ is the action output by the policy, and $R$ is the received reward. Let $(U^Z, U^X, U^A, U^R)$ be the corresponding exogenous variables that determine the values of $(Z,X,A,R)$, respectively. We adopt the three-step procedure outlined in Section~\ref{ssec:scm} to infer the counterfactual context. To ensure well-defined policies that depend on the counterfactual context, we assume that $U^X$ is uniquely determined by $x$ and $z$. With a slight abuse of notation, we denote by $U^X(\cdot)$ the corresponding mapping function. Without this assumption, this quantity will be a random variable even when conditioned on observed information, resulting in an ill-defined policy. Here we detail the counterfactual inference procedure. We begin by inferring the value of $U^X$ using $x$ and $z$, i.e., $U^X(x,z)$. Next, we perform the do-operation $do(Z=z')$ by setting the value of $Z$ to $z'$. Third, based on the inferred $U^X(x,z)$ and intervened $z'$, we compute the counterfactual context, denoted as $X^{Z\leftarrow z'}(U^X(x,z))$.  The counterfactual action $A^{Z\leftarrow z'}(U^X(x,z))$ can be calculated by applying policy $\pi$ to this counterfactual context $X^{Z\leftarrow z'}(U^X(x,z))$. We formally define CF for a policy $\pi$ in the contextual bandit framework as follows:

 \begin{definition}[Counterfactual Fairness in Contextual Bandit]\label{def:cf_policy_cb}
Given an observed context $X=x, Z=z$, a policy $\pi$ is counterfactually fair if it satisfies the following equality:
    \vspace{-0.5cm}
    \begin{equation}
        \P^{\pi}\left(A^{Z\leftarrow z'}(U^X(x,z)) = a\right) = \P^{\pi}\left(A^{Z\leftarrow z}(U^X(x,z)) = a\right),
    \vspace{-0.5cm}
    \end{equation} \label{eq:cf_def_cb}
    for any $z' \in \mathcal{Z}$ and $a \in \mathcal{A}$.
\end{definition}

\noindent To understand this definition, consider an individual with sensitive attribute $Z=z$ who receives an action $A=a$ according to policy $\pi$. CF requires that $\pi$ must assign the same action $a$ for the same individual, if in a hypothetical world their sensitive attribute $Z$ were changed to any other valid value $z' \in \mathcal{Z}, z' \neq z$.

There are two potential reasons for unfairness to happen if a policy incorporates information from both $X$ and $Z$. First, inclusion of $Z$ in the policy will make the decisions dependent on the value of $Z$ and thereby introducing potential unfairness. Second, $X$ may also have implicit information about $Z$ because of the causal arrow $Z \rightarrow X$. Even when $Z$ is excluded from being used in a policy, $X$ inherently carries information about $Z$, potentially leads to unfair decisions. We can remove the potential unfairness introduced by the impact of $Z$ on $X$ via the FLAP algorithm \citep{chen2023learning}. For a given tuple $(z_i,x_i,a_i,r_i)$ for individual $i$, we can remove the influence of $z_i$ from $x_i$ through the following procedure that produces a vector of states under the real and counterfactual worlds: $
\tilde{\bx}_i = \left\{x_i - \E_N(X \mid Z=z_i) + \E_N(X \mid Z=z)\right\}_{z \in \mathcal{Z}}.
$
\noindent This preprocessing step generates de-biased experience tuples $(\tilde{\bx}_i,a_i,r_i)_{i=1,\dots,N}$. $\tilde{\bx}_i$ represents the set of all counterfactual states corresponding to all possible values of $Z$. By construction, $\tilde{\bx}_i$ remains invariant across counterfactual values of $Z$, thereby making the resulting policy not dependent on $Z$ and counterfactually fair. One can then use the preprocessed $(\tilde{\bx}_i,a_i,r_i)_{i=1,\dots,N}$ as inputs to any off-the-shelf policy learning algorithm to learn a CF policy under contextual bandits. As we will discuss in Remark \ref{remark:sequential_rationale}, this algorithm cannot be directly applied in sequential settings for which our proposal addresses.


\subsection{Counterfactual fairness under CMDP} \label{ssec:cf_cmdp}
We are ready to generalize the definition of CF from contextual bandit to CMDP settings. Let $\bar{U}^S_t$ and $\bar{U}^R_t$ be the sequence of corresponding noise variables for $\bar{S}_t$ and $\bar{R}_t$, respectively.  The following additional assumption adapted from the contextual bandit settings is needed to ensure that the decision rule $\pi_t$ is properly defined: 

 \begin{assumption} \label{asm:unique}
     For any $t>1$, $U_t^S$ and $U_{t-1}^R$ are deterministic functions of $H_t$. With a slight abuse of notation, we denote by $U^S_t(\cdot)$ and $U^R_{t-1}(\cdot)$ the corresponding mapping functions, such that: $U^S_t = U^S_t(H_t)$, and $U^R_{t-1} = U^R_{t-1}(H_t)$.Furthermore, let $\bar{U}_t(\cdot)$ be a vector-valued function with component functions: $(U_1^S(\cdot),U_1^R(\cdot),\dots,U_{t-1}^R(\cdot),U_t^S(\cdot)).$
\end{assumption} 
Following Pearl's three-step procedure, we use $A_{t}^{Z\leftarrow z'}(\bar{U}_{t}(h_t))$ to denote the counterfactual action that would have been taken following a decision rule $\pi_t$ for an individual with history $h_t$ had their sensitive attribute been set to $z'$. With these notation and assumptions established, we can now formally define CF for a decision rule $\pi_t$ in the CMDP framework,


\begin{definition}[Counterfactual Fairness in CMDP] \label{def:cf_policy}
    Given an observed trajectory $H_t = h_t$ where $h_t = \{z,\bar{s}_t, \bar{a}_{t-1}, \bar{r}_{t-1}\}$, a decision rule $\pi_t$ is counterfactually fair at time $t$ if it satisfies the following condition:
    \vspace{-0.5cm}
    \begin{equation}
        \P^{\pi_t}\left(A_{t}^{Z\leftarrow z'}(\bar{U}_t(h_t)) = a\right) = \P^{\pi_t}\left(A_{t}^{Z\leftarrow z}(\bar{U}_t(h_t)) = a \right),
    \vspace{-0.5cm}
    \end{equation} \label{eq:cf_def}
    for any $z' \in \mathcal{Z}$ and $a \in \mathcal{A}$.
\end{definition}

\noindent Similar to Definition~\ref{def:cf_policy_cb}, CF requires that $\pi_t$ must assign the same action $a$ for the same individual with history $h_t$, if in a hypothetical world their sensitive attribute $Z$ were changed to any other valid value $z' \in \mathcal{Z}, z' \neq z$, while experiencing the same action sequence $\bar{a}_{t-1}$. A policy ${\pi} = \{\pi_t\}_{t\geq1}$ is said to satisfy CF under CMDPs if each $\pi_t$ satisfies the  Definition~\ref{def:cf_policy}.



\begin{remark}
Definition~\ref{def:cf_policy} aligns with Pearl's three-step counterfactual inference procedure \citep{pearl_causal_2016}, which we can break down as follows:
\begin{itemize}[leftmargin=*]
    \item \textbf{Abduction step}: $\bar{U}_t$ is the historical values of the background variables that follow the distribution of $P(\bar{U}_t|H_t = h_t)$ that update our knowledge of $\bar{U}_t$ given the observed trajectories $h_t$;
    \item \textbf{Action step}: We perform \textit{do}-operation: $Z$ is set to a hypothetical value $z^\prime$;
    \item \textbf{Prediction step}: Based on our updated knowledge of $\bar{U}_t$ given $h_t$, $z'$ and $\bar{a}_{t-1}$, we can predict the distribution of $A_t$ following $\pi_t$, the decision rule executed by the agent. 
\end{itemize}
\end{remark}

\section{Characterizing Counterfactually Fair Policies}\label{sec:cf_class}

In this section, we theoretically characterize the class of CF policies under CMDPs, and show the optimal CF policy is stationary when CMDPs are stationary assuming known counterfactual states and rewards. In Section \ref{sec:learning}, we will present identification assumptions and an algorithm for learning the counterfactual states and rewards from data. 

%

Given the observed history $H_t = h_t$, let $\mathcal{S}_t$ denote the set of counterfactual states at all possible values of $Z$ at time $t$. Similarly, let $\mathcal{R}_t$ denote the set of counterfactual rewards. Formally, $\mathcal{S}_t = \left\{S_{t}^{Z\leftarrow z^{(k)}}(\bar{U}_t(h_t))\right\}_{k=1,\dots,K}$, and $\mathcal{R}_t = \left\{R_{t}^{Z\leftarrow z^{(k)}}(\bar{U}_t(h_{t+1}))\right\}_{k=1,\dots,K}$. In addition, let $ \bar{\mathcal{S}}_t= \{\mathcal{S}_{t'}\}_{t'\leq t}$ and $\bar{\mathcal{R}}_t = \{\mathcal{R}_{t'}\}_{t'\leq t}$.

\begin{theorem}[Counterfactual augmentation]\label{thm:cf}
    Given observed history $H_t = h_t$ under CMDPs, $\pi_t$ satisfies CF if it admits the following functional form that
    \vspace{-0.5cm}
    \begin{equation}
    \pi_t\left(\bar{\mathcal{S}}_t,\bar{\mathcal{R}}_{t-1},\bar{a}_{t-1}\right), ~\textrm{for any}~t.\label{eq:cf_form}
    \vspace{-0.5cm}
    \end{equation}
\end{theorem}


\noindent In what follows, we will focus the class of CF policies that take $(\bar{\mathcal{S}}_t,\bar{\mathcal{R}}_{t-1},\bar{a}_{t-1})$ as input. In DTR settings, the policy that takes observables $(Z,\bar{S}_t,\bar{R}_{t-1},\bar{A}_{t-1})$ as input would achieve value optimality but violates the CF definition due to its direct use of $Z$. Our interested policy class of form (\ref{eq:cf_form}) removes direct or indirect impact of $Z$  in a causal framework, which ensures CF but potentially at the cost of reduced value due to the loss of $Z$'s information.

In RL, stationarity is often assumed for efficient policy learning \citep{sutton2018reinforcement}. Building on this, we  consider stationary CMDPs, where the system dynamics remain invariant over time. 
A history-dependent policy ${\pi}$ is a sequence of decision rules ${\pi}=\{\pi_t\}_{t\geq1}$ where each $\pi_t$ maps $(\bar{\mathcal{S}}_t, \bar{\mathcal{R}}_{t-1}, \bar{a}_{t-1})$ to a probability mass function on $\mathcal{A}$. When there exists some function $\pi^*$ such that $\pi_t(\bar{\mathcal{S}}_t, \bar{\mathcal{R}}_{t-1}, \bar{a}_{t-1}) = \pi^*(\mathcal{S}_t)$ for any $t\geq1$ almost surely, we refer to $\pi$ as a stationary policy. Let \textit{HCF}, \textit{SCF} denote the class of history-dependent and stationary CF policies, respectively. The following theorem characterizes optimal CF policies under stationary CMDPs.
  
\begin{theorem}[Optimality in stationary CMDPs]\label{thm:sr_policy}
    Under a stationary CMDP, there exists some $\pi^{opt} \in \text{SCF}$ such that 
    \[J(\pi^{opt}) = \sup_{\pi \in \text{HCF}} J(\pi),\]
    where $J(\pi) = \E_{\pi}\left[\sum_{t=0}^\infty \gamma^t R_t \right ]$ is integrated expected discounted cumulative reward with discount factor $\gamma\in (0,1)$.
\end{theorem}
\noindent Theorem~\ref{thm:sr_policy} implies that the optimal CF policy in a stationary CMDP can be found within the class of \textit{SCF}, which is much smaller than \textit{HCF}. Leveraging existing offline RL algorithms with minimal modifications, we can pool information across time to better learn a single stationary policy. 


Theorem~\ref{thm:cf} and \ref{thm:sr_policy} suggest that in stationary CMDPs, the optimal CF policy is stationary, where the decisions rely on the most recent counterfactual states at all possible values of $Z$. However, we can only observe the factual but not the counterfactual states, limiting direct application of these theoretical results. To bridge this gap, we propose a sequential data preprocessing algorithm (Section \ref{sec:learning}) that estimates these unobserved counterfactuals. This preprocessing step enables the application of existing offline RL algorithms to learn optimal CF policies using the preprocessed experience tuples.

\section{Counterfactually Fair Policy Learning}\label{sec:learning} 
In this section, we focus on the problem of learning optimal CF policies acknowledging that counterfactuals are unobserved. Our proposed approach has two steps. First, we remove sensitive attribute information from the original dataset via a sequential data preprocessing procedure (Algorithm \ref{alg:cf:1}). Second, the preprocessed dataset is used as input to any existing offline RL algorithm to learn the optimal CF policy. Similar to the single-stage setting reviewed in Section \ref{sec:contextualbandit}, a key ingredient in the first step is to accurately estimate the counterfactual states and rewards from the observed data at each time point, for which we introduce the following assumption:



\begin{assumption}[Additivity]\label{asm:additive}
    For all time $t\geq 1$, the exogenous variables $U^{S}_{t}$ and $U^{R}_{t}$ are additive to $S_t$ and $R_t$, respectively.
\end{assumption}

\noindent 
Assumption~\ref{asm:additive}, which enables the estimation of exogenous variables $\{U^S_t\}_{t\geq1}$ and $\{U^R_t\}_{t\geq1}$, allows us to identify and estimate the counterfactual states and rewards using observed data. This assumption is related to \textit{level $3$} assumption in \citet{kusner2017counterfactual}'s work, which maximizes the information the policy learner can use. \citet{kusner2017counterfactual} also introduced \textit{level $1$} and \textit{$2$} assumptions, which are more flexible than this additivity assumption.

With all the necessary assumptions established, we now present Algorithm~\ref{alg:cf:1} of the proposed sequential data preprocessing procedure. We use $\hat{s}_{it}(z')$ and $\hat{r}_{it-1}(z')$ to represent the estimated values of $S_t^{Z\leftarrow z'}(\bar{U}_t(H_t))$ and $R_{t-1}^{Z\leftarrow z'}(\bar{U}_t(H_t))$ for individual $i$, respectively. The algorithm's core strategy is to leverage observed data to estimate counterfactual states and rewards under the additive assumption. These estimates enable us to construct preprocessed experience tuples from the original data, which can then be used to train optimal policies that satisfy CF requirements. Notably, these preprocessed experience tuples naturally form a MDP, as shown in the proof of Theorem~\ref{thm:sr_policy}, allowing us to apply existing RL algorithms directly. 
It is important to note that the rationale for preprocessing $s_{it}$ and $r_{it}$ differs. According to Theorem~\ref{thm:cf}, pre-processing $s_{it}$ ensures that the learned policy achieves CF and there is no requirement on reward $r_{it}$. One can use preprocessed $\hat{s}_{it}$ and observed $r_{it}$ to learn a CF policy. However, the purpose of preprocessing $r_{it}$ is to ensure that the learned policy maximizes the cumulative discounted reward under stationary CMDPs based on Theorem~\ref{thm:sr_policy}.

\begin{algorithm}[ht!]
    \begin{algorithmic}[1] \onehalfspacing
        \Require Original data $\mathcal{D}=\{(s_{it},z_i,a_{it},r_{it}):i=1,\dots,N;t=1,\dots,T\}$.
        \State Fit the mean of transition kernel $\hat{\mu}(s,a,z)$ by minimizing mean squared error (MSE) on $\mathcal{D}$.
        \State Estimate $\hat{\E}(S_1|Z=z)$ and $\hat{P}(Z=z)$ $\forall z' \in \mathcal{Z}$  by the empirical means.
        \For{$i=1,\dots,N$}
        \State Calculate $\hat{s}_{i1}^{z'} = s_{i1} - \hat{\E}(S_1|Z=z) + \hat{\E}(S_1|Z=z'), \forall z' \in \mathcal{Z}$.
        \State Set $\tilde{s}_{i1} = [\hat{s}_{i1}^{z^{(1)}},\dots,\hat{s}_{i1}^{z^{(K)}}]^\top$.
        \For{$t=2,\dots,T$}
			\State $[\hat{s}_{it}^{z'},\hat{r}_{i,t-1}^{z'}]^\top = [s_{it},r_{i,t-1}]^\top - \hat{\mu}(s_{i,t-1},a_{i,t-1},z_i) + \hat{\mu}(\hat{s}_{i,t-1}^{z'},a_{i,t-1},z'), \forall z' \in \mathcal{Z}$.
            \State $\tilde{s}_{it} = [\hat{s}_{it}^{z^{(1)}},\dots,\hat{s}_{it}^{z^{(K)}}]^\top$, 
            \State $\tilde{r}_{i,t-1} = \sum_{k=1}^K \hat{P}(Z=z^{(k)}) \hat{r}_{i,t-1}^{z^{(k)}}.$
        \EndFor
        \EndFor
        \Ensure  Preprocessed experience tuples $\{(\tilde{s}_{it},a_{it},\tilde{r}_{it}):i=1,\dots,N;t=1,\dots,T\}$.
    \end{algorithmic}
    \caption{Proposed sequential data preprocessing}
    \label{alg:cf:1}
\end{algorithm}

\begin{remark}\label{remark:sequential_rationale}
    Algorithm~\ref{alg:cf:1} generalizes the data preprocessing algorithm in \citet{chen2023learning}, which shares the idea of estimating the counterfactual states through preprocessing. Although their approach can be extended to single-stage contextual bandit settings, it falls short in multi-stage CMDP settings. The challenge is that the information of $Z$ is embedded in the states $S_t$ for every time point $t\geq1$. To infer the counterfactual state $S_t^{Z\leftarrow z'}(\bar{U}_t(H_t))$ at time $t$, it is necessary to first compute the value of preceding counterfactual state $S_{t-1}^{Z\leftarrow z'}(\bar{U}_{t-1}(H_{t-1}))$ (Line 7, Algorithm~\ref{alg:cf:1}). This sequential dependency motivates the proposed sequential data preprocessing procedure, where counterfactual states are inferred from $t=1$ to $T$.
\end{remark}

\subsection{Theoretical analysis}
\label{sec:theory_policy}

In this section, we establish the theoretical results on the regret, which measures the difference between the expected cumulative reward under the optimal policy and that under the estimated policy, and unfairness control of the optimal policy learned using Algorithm~\ref{alg:cf:1} and fitted Q iteration \citep[FQI, ][]{riedmiller2005neural} in tandem. FQI is widely used policy learning algorithm in RL, where the procedure is detailed in Algorithm~\ref{alg:cf:2}.  Unlike traditional FQI analyses that work on observed states and rewards, our algorithm requires estimating the counterfactual states and rewards before applying FQI. Therefore, this section presents a novel FQI analysis tailored to the scenario where both states and rewards are estimated. For ease of presentation, we use $\hat{s}$ and $\hat{r}$ to represent estimated quantities in theoretical analysis.

\begin{algorithm}[ht!]
    \begin{algorithmic}[1] \onehalfspacing
        \Require Dataset $\mathcal{D}=\{(s_{i},a_{i},r_{i}, s_{i}^\prime):i=1,\dots,n\}$, function class $\mathcal{F}$.
        \State Initialize $\hat{f}_0 \in \mathcal{F}$ randomly.
        \For{$b=1,\dots,B$}
            \State Compute target $y_i = r_i + \gamma \max_a \hat{f}_{b-1}(s_i^\prime,a)$.
            \State Build training dataset $D_b = \{s_i, a_i, y_i\}_{i=1,\dots,n}$.
            \State Solve a supervised learning problem:
            \vspace{-0.3cm}
            \[\hat{f}_b = \arg \min_{f \in \mathcal{F}} \frac{1}{n} \sum_{i=1}^n \left( f(s_i,a_i) - y_i \right)^2 \vspace{-0.3cm}\]
        \EndFor
         \Ensure Estimated optimal Q function $\hat{f}_B$.
    \end{algorithmic}
    \caption{Fitted Q iteration \citep{riedmiller2005neural}.}
    \label{alg:cf:2}
\end{algorithm}

Here we briefly introduce the assumptions used in the analysis. First, our results are based on the FQI algorithm on the hypothesis class $\mathcal{F}$ that is linear in $d$-dimensional features $\phi(s,a)$, with bounded coefficients $w \in \mathbb{R}^d$ (Assumption~\ref{asm:hypothesis}). Second, we assume that $\mathcal{F}$ is closed under the Bellman optimality operator (Assumption~\ref{asm:completeness}), which is known as the Bellman completeness assumption \citep{chen2019information}. Third, we assume sufficient feature coverage in the dataset (Assumption~\ref{asm:coverage}). More specifically, we require that the minimum eigenvalue of $\E_{s,a \sim \mu_b}[\phi(s,a) \phi(s,a)^T])$ is greater than $\lambda_0$ where $\mu_b$ is the data distribution. This assumption is commonly required to guarantee the convergence of FQI estimators \citep{wang2020statistical,hu2024fast}. Fourth, we assume that each feature $\phi_i(s,a)$ is a $L-$Lipschitz continuous function for any $i=1, \ldots, d$ and $a$ (Assumption~\ref{asm:lipschitz}). Fifth, we assume that the reward are bounded, i.e., $|r|\leq R_{\max}$ (Assumption~\ref{asm:bounded_reward}). We denote $n$ as the number of experience tuples in the dataset, i.e., $n = NT$.

\begin{theorem}[Regret Bound] \label{thm:regret}
 Let $\epsilon = \max_{s}\|\hat{s}-s\|_\infty + \max_{r}|\hat{r}-r|$. Suppose Assumptions~\ref{asm:hypothesis}-\ref{asm:bounded_reward} holds.  With probability at least $1 - n^{-\kappa} - d\exp(-n\lambda_0/8)$ the regret of the optimal policy estimated using Algorithm~\ref{alg:cf:1} and FQI is upper bounded by
 \vspace{-0.5cm}
    \begin{equation}
        \frac{ C dLR_{\max}\epsilon}{\lambda_0(\lambda_0-4dL\epsilon)(1-\gamma)^3} + \frac{ C dR_{\max}\kappa\log(n)}{\lambda_0\sqrt{n}(1-\gamma)^3} + \frac{\gamma^B R_{\max}}{(1-\gamma)^2},\label{eq:xi}
         \vspace{-0.5cm}
    \end{equation}
for any $\kappa > 0$, some positive constant $C>0$ and $B$ is the number of FQI iterations. 
\end{theorem}

Theorem~\ref{thm:regret} implies that the regret bound is composed of three terms: (i) The first term is proportional to $\epsilon$, which measures the \textit{estimation error of counterfactual states and rewards}; (ii) The second term is directly linked to the \textit{one-step FQI regression error}, which approaches zero as $n$ increases; (iii) The third term characterizes the \textit{initialization bias}, 
which approaches zero exponentially fast with respect to the number of FQI iterations $B$. 

Meanwhile, these error terms are also dependent upon some other factors, such as the feature dimension $d$, the Lipschitz constant $L$, the reward upper bound $R_{\max}$, the minimum eigenvalue $\lambda_0$ and the $(1-\gamma)^{-1}$-term which has a similar interpretation to the horizon in episodic tasks. Their dependencies align with existing findings in the literature \citep[see e.g.,][]{chen2019information,hu2024fast}. 


To analyze unfairness control of the learned policy, we introduce an additional margin-type assumption, as detailed in Assumption~\ref{asm:margin} in the supplementary materials. This assumption is often imposed in RL literature \citep{qian2011performance,shi2022statistical,hu2024fast}. Let $\xi$ be the FQI error bound (Equation~(\ref{eq:fqi_error_bound}) in the supplementary materials) and $\hat{\pi}$ denote the learned optimal policy using Algorithm~\ref{alg:cf:1} and \ref{alg:cf:2}.

\begin{theorem}[Unfairness Control] \label{thm:cf_metric}
Under Assumptions~\ref{asm:hypothesis}- \ref{asm:bounded_reward} and \ref{asm:margin} in the supplementary materials, suppose $\epsilon = \max_s\|\hat{s}-s\|_\infty + \max_r|\hat{r}-r|$, for any $\kappa > 0$, with probability at least $1 - n^{-\kappa} - d\exp(-n\lambda_0 / 8)$, the absolute difference in action distributions for $\hat{\pi}$ using under two counterfactual worlds ($z'$ and $z''$) is no more than $2^{\alpha+1} \xi^\alpha +2^\alpha (L\epsilon)^\alpha$. 
\end{theorem}


Theorem~\ref{thm:cf_metric} characterizes the difference in action distributions when the input counterfactual states are estimated under two worlds with distinct values of $Z$. The analysis implies that the unfairness is influenced by two terms: (i) The first term is related to the FQI estimation error. As this error goes smaller, the unfairness also decreases. 
(ii) The second term is related to the estimation error of counterfactual states. As this estimation error becomes smaller, the estimated counterfactual states under different  $z$'s will be similar, resulting in generating similar action distributions.

\section{Numerical experiments} \label{sec:num_exp}

In this section, we evaluate the performance of our approach using synthetic and semi-synthetic datasets that mirror the distributional characteristics of the PowerED study data. Our assessment focused on two key metrics: (1) the value attained by the learned policy, and (2) the degree of counterfactual unfairness. The latter is operationalized as a measure of how actions differ between the observed world and a hypothetical world where only the sensitive attribute is altered. 

\paragraph{Baselines} Five baselines are considered: 1) \textit{Full}, a standard policy that uses all variables, including the sensitive attribute and other state variables, to make decisions; 2) \textit{Unaware}, a policy that uses all variables except the sensitive attribute to make decisions; and 3) \textit{Oracle}, an idealized policy that used concatenations of counterfactual states where the counterfactual states and rewards are assumed known. For reference, we also include 4) \textit{Random} policy that selects actions randomly and 5) \textit{Behavior} policy that is used to collect data. By definition, \textit{Random} policies naturally satisfy CF but may not achieve high values.

\paragraph{Fairness metric} To measure deviation from counterfactual fairness, we introduce the following CF metric, adapted from previous work \citep{chen2023learning,wang2022adjusting,wu2019counterfactual},
\begin{align}\label{eq:cf_metric}
    \max_{z',z \in \mathcal{Z}}\frac{1}{NT} \sum_{i=1}^N \sum_{t=1}^{T} \mathbf{I} \left( A_{t}^{Z\leftarrow z'}(\bar{U}_{t}(h_{it})) \neq  A_{t}^{Z\leftarrow z}(\bar{U}_{t}(h_{it}))\right). 
\end{align}
This metric calculates the maximum discrepancy between the average discordance rate between actions in the factual and counterfactual worlds across all the time points for any given pair of distinct sensitive attribute values, $(z,z')$ where $z,z' \in \mathcal{Z}, z \neq z'$. A lower value indicates that the policy is fairer. The metric is bounded between 0 and 1, with 0 representing perfect fairness and 1 indicating complete unfairness.

\paragraph{Deployment of our approach:} To deploy the learned policy using Algorithm~\ref{alg:cf:1} and \ref{alg:cf:2}, counterfactual states need to be sequentially estimated. Specifically, before making a decision at time $t$ using $\tilde{x}_t$, the policy needs to first use the stored $\tilde{x}_{t-1}$ and observed $x_t$ to estimate the value of $\tilde{x}_t$. Therefore, the learned policy requires a memory buffer to store the counterfactual states at previous time point during deployment.

\subsection{Synthetic data experiments}
We consider both linear and non-linear transition kernels settings in this experiment. We consider two goals: 1) validate that the level of counterfactual unfairness decreases as the sample size increases for our approach, and 2) investigate how the strength of a sensitive attribute's impact on state and reward affects unfairness. For each setting, we fix the length of horizon $T=10$. Let $\delta$ denote the strength of impact of sensitive attribute $Z$ on states $\{S_t\}_{t \geq 1}$ and rewards $\{R_t\}_{t \geq 1}$. For the first goal, we vary the sample size $N$ in $\{100,200,500,1000,2000\}$ and fix $\delta$ to be $1$. For the second goal, we fix the sample size $N=1000$ and vary $\delta$ in $\{0.0,0.5,1.0,1.5,2.0\}$. 
The CF metric is calculated by comparing the action distributions generated by the policy in the observed and corresponding counterfactual world. The cumulative discounted reward is calculated using the observed reward collected from the trajectories. We use $N=10000$ and $T=20$ to calculate each metric. 

\begin{figure}[t]
    \centering
    \includegraphics[width=0.95\textwidth]{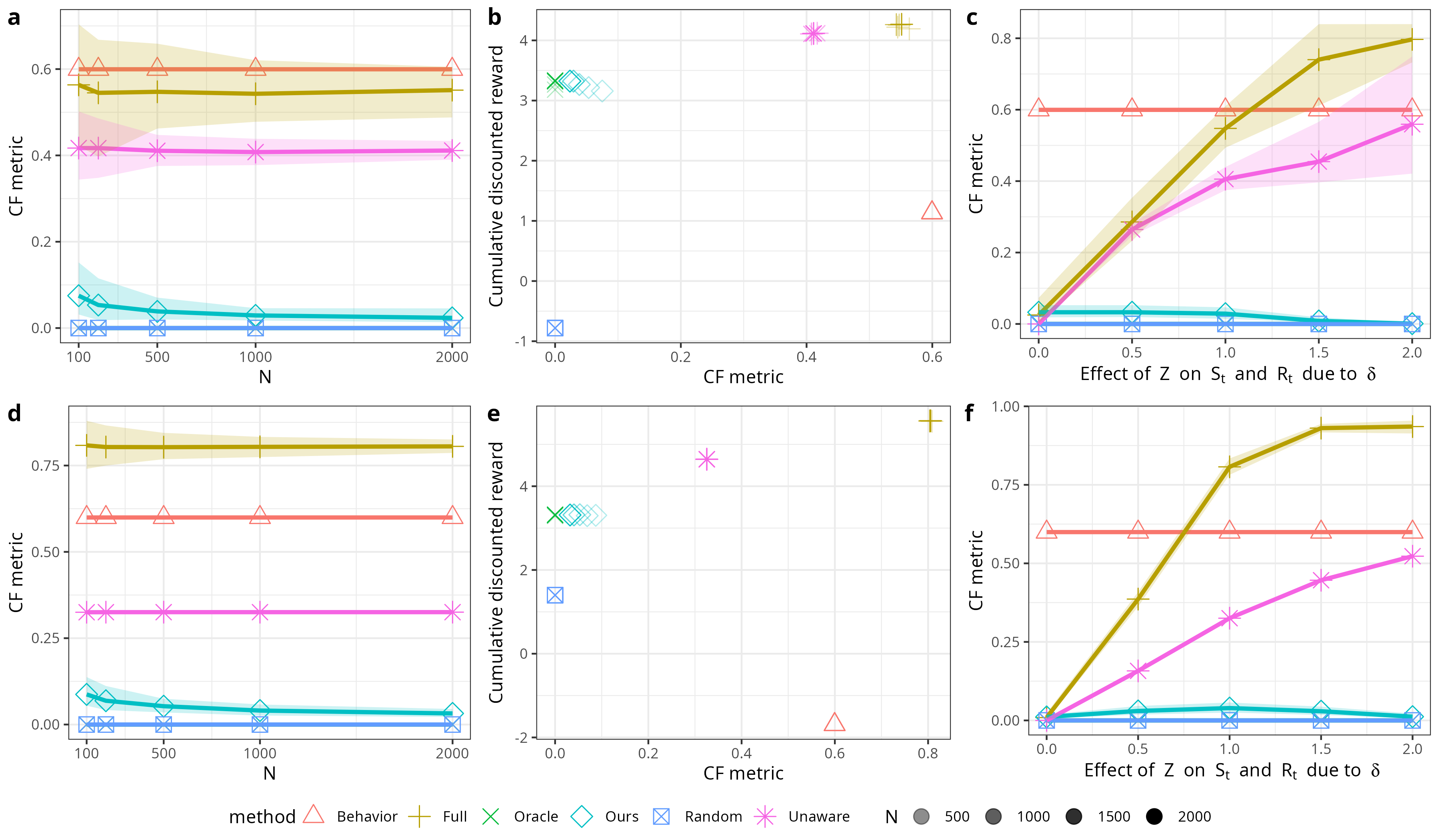}
    \caption{Comparisons under the linear setting (top row) and non-linear setting (bottom role): a,d) CF metric versus sample size, b,e) cumulative discounted reward versus CF metric (multiple sample sizes) with $\delta=1$, c,f) CF metric versus $\delta$. All the results are aggregated over 100 random seeds. The shaded area represents the $95\%$ CI.}\label{fig:sim_all}
\end{figure}

Figure~\ref{fig:sim_all}(a,b,d,e) present the results for the first goal. As expected, both \textit{Random} and \textit{Oracle} policies achieves prefect CF. The \textit{Full} and \textit{Unaware} policies exhibits high unfairness levels, while our proposed approach achieves lower CF metric. Additionally, the CF metric of our approach decreases with increasing sample size $N$, validating the consistency. In terms of cumulative discounted reward, the \textit{Full} policy achieves the greatest total reward due to its access to all state information. The other three approaches have lower reward due to the loss of state information, indicating a fairness-reward trade-off \citep{dutta2020there,wick2019unlocking}. To control the degree of unfairness, a weighted combination of the \textit{Full} policy and our proposed policy could be considered. Greater weight assigned to our proposed method prioritizes fairness, and conversely, greater weight on the \textit{Full} policy prioritizes reward maximization.

Figure~\ref{fig:sim_all}(c,f) show the results for the second goal. We observe that the unfairness of the \textit{Full} and \textit{Unaware} policies increases with increasing $\delta$, while our proposed approach effectively controls unfairness, indicating that our approach can effectively remove the information of sensitive attribute from the state variables and learn CF policy at varying vulnerabilities to unfairness.

\subsection{Semi-synthetic data }\label{sec:semi}

\begin{figure}
    \centering
    \includegraphics[width=0.95\textwidth]{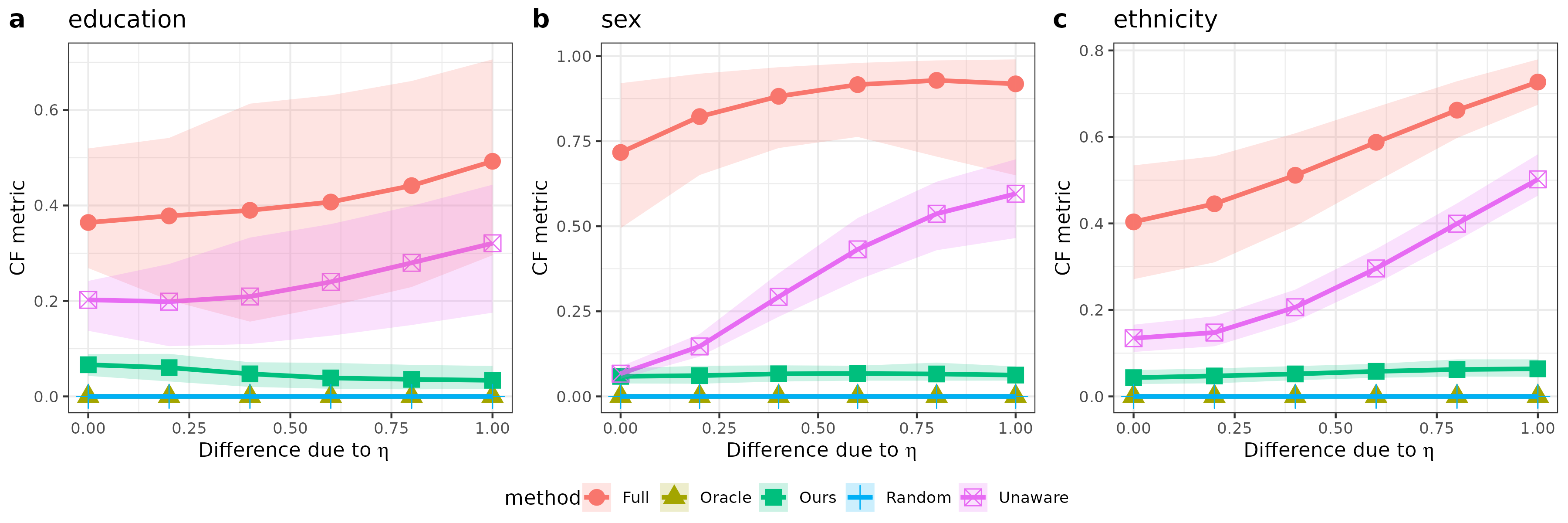}
    \caption{Semi-synthetic experiments: CF metric of different approaches versus $\eta$ for different sensitive attributes: education, sex, ethnicity. All the results are aggregated over 100 random seeds. The shaded area is the $95\%$ CI.}\label{fig:semi}
\end{figure}

To bridge the gap between theoretical results and practical applicability, we conduct real-world data-based simulations to evaluate the performance of different approaches under more realistic conditions. These simulations were designed to mimic the motivating PowerED study \citep{piette2023automatically} with respect to transition dynamics. We select education, sex and ethnicity as three different potentially sensitive attributes in our simulations. The state variables include weekly pain intensity score and pain interference score - two commonly used metrics to evaluate progress in pain management programs. The reward is the \texttt{7 - weekly self-reported opioid medication risk score}, defined in detail in \citet{piette2023automatically}. We simulate $100$ different datasets with $N=1000$ and $T=12$ using the neural network-based generative models learned from the real dataset.  To investigate how the strength of the sensitive attribute's state variables affects unfairness, we incorporated a strategy similar to the one used in the synthetic data experiments.  We varied the effect magnitude of the sensitive attribute on the state variables by adding a constant value $\eta$ to the state variables for one value of sensitive attribute. The results are shown in Figure~\ref{fig:semi}. We observe that the \textit{Full} policy has high levels of counterfactual unfairness across all scenarios. The \textit{Unaware} policy demonstrates increasing unfairness as the effect of the sensitive attribute on the state variables $\eta$ increases. This suggests that even after excluding the sensitive attribute, the state variables might still contain residual information of the sensitive attribute, resulting in unfairness. Our proposed approach achieves lower counterfactual unfairness compared with the other two methods. The observed unfairness in our method is primarily due to the approximation error of the counterfactual state when compared with the perfect fairness for the \textit{Oracle} policy. The results are in line with our theoretical ﬁndings.

\section{Application to Reducing Opioid Misuse Behaviors}\label{sec:real_data}

In this section, we apply the proposed algorithm to a real-world digital health dataset named PowerED study, as described in Section~\ref{sec:intro}. The data comprise 207 patients over 12 weeks. We selected education, age, gender, and ethnicity as the potentially sensitive attributes, and weekly pain and pain interference scores as the state variables. The reward is defined as \texttt{$7$ - weekly self-reported opioid medication risk score}, which was measured by items during weekly monitoring calls.

We compare our approach to three baselines: \textit{Full}, \textit{Random} and \textit{Unaware}, as detailed in Section~\ref{sec:num_exp}. CF metric and cumulative discounted reward are evaluated. To compute the CF metric, we need to assess counterfactual actions under different values of $Z$, while we only observe a single $z$ in the dataset. Therefore, we train a generative model, similar to the one presented in Section~\ref{sec:semi}, to generate counterfactual actions, enabling the calculation of the CF metric. We use fitted Q evaluation \citep{le2019batch} to estimate the value for each policy. The dataset is split into training and validation sets using an $80:20$ ratio, where the training set is used for policy learning and validation set is used for metric calculations. The results are aggregated for $10$ different seeds.

As summarized in Table~\ref{table:real_data},  \textit{Random} policy achieves the perfect fairness but a lower value, as it entirely disregards state information for decision-making. \textit{Full} policy demonstrates the highest value for most variables, however, it has the highest level of unfairness, as incorporating sensitive attributes into the decision-making process can inherently lead to unfair decisions. Our approach achieves the lowest unfairness for all four sensitive attributes compared to other methods. While our approach did not achieve perfect fairness, we believe that the approximation errors associated with estimating counterfactual states and rewards are likely to be relatively higher. Supplemental Figure~\ref{fig:real_action_percentage} illustrates the favorability of actions of different policies towards different sensitive groups. It can be seen that the \textit{Full} policy tends to favor younger, more educated, non-Hispanic, and male patients by giving them a higher percentage of treatments compared to other sensitive groups. Our approach mitigates this bias and achieves lower unfairness. Our approach greatly enhanced fair access to human counselors without significantly compromising population-level benefit.

\begin{table}[htp]
\centering

\begin{tabular}{cccccc}
\hline
Metric                 & Method  & Education            & Age                  & Sex                  & Ethnicity            \\ \hline
\multirow{4}{*}{Unfairness}    & Full    & 0.44 (0.14)          & 0.59 (0.15)          & 0.61 (0.15)          & 0.39 (0.13)          \\
                       & Random  & 0.00 (0.00)          & 0.00 (0.00)          & 0.00 (0.00)          & 0.00 (0.00)          \\
                       & Unaware & 0.10 (0.03)          & 0.10 (0.02)          & 0.08 (0.03) & 0.21 (0.05)          \\
                       & Ours    & \textbf{0.06 (0.02)} & \textbf{0.08 (0.02)} & \textbf{0.07 (0.02)}          & \textbf{0.16 (0.03)} \\ \hline
\multirow{4}{*}{Value} & Full    & 57.09 (0.31)         & 57.29 (0.30)         & 57.20 (0.39)         & 56.87 (0.33)         \\
                       & Random  & 56.61 (0.22)         & 56.66 (0.27)         & 56.53 (0.27)         & 56.54 (0.39)         \\
                       & Unaware & 57.01 (0.18)         & 57.21 (0.29)         & 56.96 (0.32)         & 57.00 (0.31)         \\
                       & Ours    & 57.05 (0.30)         & 57.11 (0.28)         & 56.95 (0.51)         & 56.93 (0.48)         \\ \hline
\end{tabular}
\caption{Unfairness metric (the lower the better) and value (the higher the better) for different approaches in the real data analysis.} \label{table:real_data}
\end{table}

\section{Discussion} \label{sec:conclusion}

In this paper, we have studied counterfactual fairness in the novel context of RL. We provide a general framework for defining CF in multi-stage settings and theoretically characterize the class of CF policies. We also prove that the optimal CF policy  is  stationary under stationary CMDPs, which greatly simplifies policy learning. Methodologically, we develop a novel sequential data preprocessing algorithm designed to mitigate bias in multi-stage settings and to  produce preprocessed experience tuples to enable learning CF policies by leveraging existing RL methods. We also establish theoretical guarantees on value optimality and unfairness control. Empirical results based on numerical experiments corroborate with our theory. We provide some additional discussions.

\paragraph{Definition~\ref{def:cf_policy}} Definition~\ref{def:cf_policy} extends path-dependent CF \citep{kusner2017counterfactual} to the CMDP setting by fixing the historical action sequence $\bar{A}_{t-1}$ to the observed sequence $\bar{a}_{t-1}$ rather than allowing the past actions to change after switching the value of $Z$. A direct extension of single-stage CF \citep{kusner2017counterfactual} to CMDP without fixing $\bar{A}_{t-1}$ to $\bar{a}_{t-1}$ would be problematic, as the resulting CF definition would depend on the behavior policy. A further extension to Definition \ref{def:cf_policy} would require action distribution invariance for arbitrary historical action sequences $\bar{a}_{t-1}^\prime \in \mathcal{A}^{t-1}$ that could have but not actually experienced by the individual in the past (Definition~\ref{def:cf_policy_strict}, Section~\ref{fair_rl:apdx:strict_cf}  in the supplementary materials). The extension requires correcting historical unfair decisions the individual have received before an action at time $t$, while Definition \ref{def:cf_policy} solely focuses on mitigating decision unfairness at and later than time $t$ assuming past actions $\bar{A}_{t-1}$ cannot be undone. Definition~\ref{def:cf_policy_strict} is therefore stricter than Definition~\ref{def:cf_policy} and will result in a lower optimal value. 

\paragraph{Class of CF Policies} We focused on the class of CF policies that takes $(\bar{\mathcal{S}}_t,\bar{\mathcal{R}}_{t-1},\bar{a}_{t-1})$ as input. Do there exist CF policies outside this class that can achieve higher value? Consider the three policy classes below with increasing restrictiveness. The first class is $\Pi_1 = \{\pi_t:\pi_t(Z,\bar{S}_t,\bar{R}_{t-1},\bar{A}_{t-1})\}$. DTR \citep{murphy2003optimal} ensures that $\Pi_1$ contains value-optimal policies, however, direct use of $Z$ violates the CF definition. The second class, $\Pi_2 = \{\pi_t:\pi_t(\bar{\mathcal{S}}_t,\bar{\mathcal{R}}_{t-1},\bar{A}_{t-1})\}$, removes the information of $Z$ from the policy inputs, which ensures CF but potentially at the cost of reduced value compared to $\Pi_1$. The third class, $\Pi_3$, extends beyond $\Pi_2$ by applying do-calculus not only to $Z$ but also to $\bar{A}_{t-1}$ (to any action sequence). While policies in $\Pi_3$ satisfy CF, they achieve lower value than those in $\Pi_2$ because the information of the observed $\bar{A}_{t-1}$ is removed from policy input.

\paragraph{Additivity Assumption} The additive assumption (Assumption~\ref{asm:additive}) can be restrictive in real-world settings. However, it is important to acknowledge that causal inference, especially when dealing with counterfactuals, often relies on strong assumptions \citep{dawid2000causal,pearl2009causality}. Relaxing these assumptions may be possible with more flexible approaches, such as variational autoencoder \citep{louizos2017causal} and adversarial training \citep{melnychuk2022causal}. We leave this topic for future research.


\section*{Supplementary Materials}

Supplementary Materials contain a glossary of notation, proofs, supporting figures, tables, longer derivations, and additional results. The following sections are included: \\
\textbf{A: Generalization of Definition \ref{def:cf_policy}.} \\
\textbf{B-E: Proofs of Theorems \ref{thm:cf} \ref{thm:sr_policy}, \ref{thm:regret},  \ref{thm:cf_metric}} \\
\textbf{F: Details on numerical experiments in Section \ref{sec:num_exp}}\\
\textbf{G: More details on real data analysis in Section \ref{sec:real_data}}\\
Code to reproduce simulations and data analysis will be made available online at github.com.

\bibliographystyle{agsm}
\bibliography{CFRL_main}
\makeatletter

\makeatother
\end{document}


\def\spacingset#1{\renewcommand{\baselinestretch}%
{#1}\small\normalsize} \spacingset{1}


\if1\blind
{
  \title{\bf Suppplementary Materials for ``Counterfactually Fair Reinforcement Learning via Sequential Data Preprocessing''}
  \author{Author 1\thanks{
    The authors gratefully acknowledge \textit{please remember to list all relevant funding sources in the unblinded version}}\hspace{.2cm}\\
    Department of YYY, University of XXX\\
    and \\
    Author 2 \\
    Department of ZZZ, University of WWW}
} \fi

\if0\blind
{
  \bigskip
  \bigskip
  \bigskip
    \title{\bf Suppplementary Materials for ``Counterfactually Fair Reinforcement Learning via Sequential Data Preprocessing''}
  \medskip
} \fi
\date{}
\maketitle
The supplementary materials contain a glossary of technical notation,  a technical extension, proofs of theorems, and additional details about simulation and data analysis.

\tableofcontents

\renewcommand{\thesection}{\Alph{section}}
\renewcommand\thefigure{\thesection.\arabic{figure}}    
\renewcommand\thetable{\thesection.\arabic{table}}    
\numberwithin{equation}{section}
\makeatletter 
\renewcommand\theequation{\thesection.\arabic{equation}}    
\newcommand{\section@cntformat}{Supplement \thesection:\ }
\makeatother

\renewcommand{\thedefinition}{\Alph{section}\arabic{definition}}
\renewcommand{\theassumption}{\Alph{section}\arabic{assumption}}

\newpage 
We first provide a glossary for key notation used in the Main Paper. 
\bgroup
\def\arraystretch{1.5}
\begin{table}[H]
    \centering
    \begin{tabular}[\textwidth]{p{2.5cm} p{8cm}}
      \toprule
	\centering \textbf{Symbol} & \textbf{Description} \\
      \midrule
	\centering $\bar{S}_t,\bar{A}_t,\bar{R}_t$    &    historical states, actions and rewards up to time $t$.         \\ 
	\centering $\bar{U}^S_t,\bar{U}^A_t,\bar{U}^R_t$   &    historical exogenous variables for states, actions and rewards up to time $t$.         \\ 
	\centering $H_t$  &  $\{Z,\bar{S}_t, \bar{A}_{t-1}, \bar{R}_{t-1}\}$, complete history up to time $t$     \\ 
	 \centering $\bar{U}_t$ & $\{\bar{U}^R_{t-1},\bar{U}^S_t\}$        \\ 
	 \centering $R_{t-1}^{Z\leftarrow z'}(\bar{U}_{t}(h_t))$, $S_{t}^{Z\leftarrow z'}(\bar{U}_{t}(h_t))$, $A_{t}^{Z\leftarrow z'}(\bar{U}_{t}(h_t))$ & the counterfactual reward, state, action that would have been for an individual with history $h_t$ had their sensitive attribute been $z'$, while experiencing \textbf{the observed action sequence $\bar{a}_{t-1}$} and following $\pi_t$  \\ 
	 \centering $\mathcal{S}_t$ &  $[S_{t}^{Z\leftarrow z^{(k)}}(\bar{U}_t(h_t))]_{k=1,\dots,K}$ \\ 
	 \centering $\mathcal{R}_{t-1}$ & $[R_{t-1}^{Z\leftarrow z^{(k)}}(\bar{U}_t(h_{t}))]_{k=1,\dots,K}$ \\ 
       \bottomrule
    \end{tabular}
\end{table}
\egroup

\section{Generalization of Definition~\ref{def:cf_policy}} \label{fair_rl:apdx:strict_cf}
\begin{definition}[Strict Counterfactual Fairness in CMDP] \label{def:cf_policy_strict}
    Given an observed trajectory $H_t = h_t$, a decision rule $\pi_t$ is strictly counterfactually fair at time $t$ if it satisfies the following condition:
    \begin{align}
        \P^{\pi_t}\left(A_{t}^{Z\leftarrow z',\bar{A}_{t-1} \leftarrow \bar{a}_{t-1}^\prime}(\bar{U}_t(h_t)) = a \right) = \P^{\pi_t}\left(A_{t}^{Z\leftarrow z,\bar{A}_{t-1} \leftarrow \bar{a}_{t-1}}(\bar{U}_t(h_t)) = a \right),
    \end{align}
    for any $z' \in \mathcal{Z}, \bar{a}_{t-1}^\prime \in \mathcal{A}^{t-1}$ and $a \in \mathcal{A}$.
\end{definition}
\noindent Here the definition is stricter than Definition~\ref{def:cf_policy} because it requires action distribution invariance for arbitrary historical action sequences $\bar{a}_{t-1}^\prime \in \mathcal{A}^{t-1}$ that could have but not actually experienced by the individual in the past.

\section{Proof of Theorem~\ref{thm:cf}} \label{apdx:thm1}

\begin{proof}
    If we can show that $\bar{\mathcal{S}}_t$, $\bar{\mathcal{R}}_t$ and $\bar{a}_{t-1}$ are invariant with respect to the counterfactual values of $Z$, and given $\pi_t$ is a function of these quantities, then Equation~(\ref{eq:cf_form}) is invariant to different counterfactual values of $Z$. \textbf{First}, since we fix $\bar{A}_{t-1}$ to be the observed action sequence $\bar{a}_{t-1}$ for all possible values of $Z$, $\bar{a}_{t-1}$ is invariant to the counterfactual values of $Z$. \textbf{Second}, $\mathcal{S}_{t}$ consists of the counterfactual states for all possible values of $Z$. Altering the value of $Z$ does not affect the distribution of $\mathcal{S}_{t}$. Since $\bar{\mathcal{S}}_{t}$ is a collection $\mathcal{S}_{t'}$ for $t' \leq t$, its distribution of $\bar{\mathcal{S}}_{t}$ remains invariant to counterfactual values of $Z$. Similarly, we can show that $\bar{\mathcal{R}}_t$ is invariant to any counterfactual values of $Z$. As $\bar{\mathcal{S}}_t$, $\bar{\mathcal{R}}_t$ and $\bar{a}_{t-1}$ are invariant with respect to the counterfactual values of $Z$, any decision rule $\pi_t$ satisfies Equation~(\ref{eq:cf_form}) is counterfactually fair.
\end{proof}

\section{Proof of Theorem~\ref{thm:sr_policy}} \label{apdx:thm2}
\begin{proof}
    We begin by introducing the following notations
    \begin{align*}
        \mathbb{S}_t^k &= S_{t}^{Z\leftarrow z^{(k)}}\left(\bar{U}_t\left(H_t\right)\right), \\
        \mathbb{R}_t^k &= R_{t}^{Z\leftarrow z^{(k)}}\left(\bar{U}_t\left(H_t\right)\right), \\
        \mathbb{S}_t &= \left\{\mathbb{S}_t^k\right\}_{k=1,\dots,K}, \\ 
        \mathbb{R}_t &= \left\{\mathbb{R}_t^k\right\}_{k=1,\dots,K}.
    \end{align*}
     For simplicity, we include $\mathbb{R}_t^k$ in the set of $\mathbb{S}_{t+1}^k$. The following conclusions automatically holds if we include $U^R_t$ in the set of $U^S_{t+1}$. Note that unlike $\mathcal{S}_{t}^k$, we do not enforce $\bar{A}_{t-1} \leftarrow \bar{a}_{t-1}$ on $\mathbb{S}_{t}^k$. The action sequence $\bar{a}_{t-1}$ can be arbitrary, however, $\mathbb{S}_{t}^k$ share the same $\bar{a}_{t-1}$ for different $k$ due to Pearl's three-step procedure for counterfactual inference \citep{pearl_causal_2016}. According to Assumption~\ref{asm:markov}, Figure~\ref{fig:dag_thm2} depicts the relationship between $\mathbb{S}_t^k$, $A_t$ and $U_{X_t}$.
    
    \begin{figure}[H]
        \centering
        \includegraphics[width=0.8\linewidth]{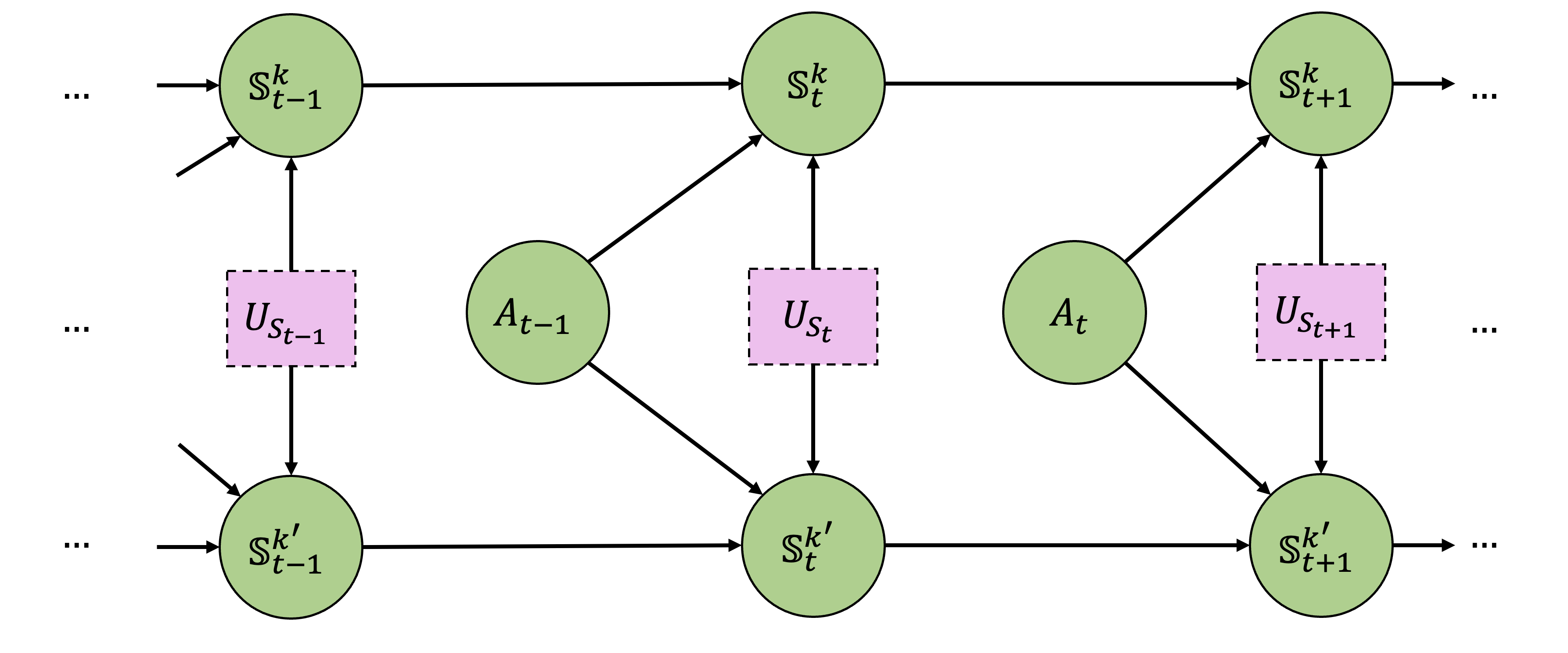}
        \caption{Causal diagram of counterfactual states and actions.}
        \label{fig:dag_thm2}
    \end{figure}

    \noindent We have following conclusions via d-separation: for any $k$
    \begin{align*}
        &\mathbb{S}_{t+1}^k \ind \{\mathbb{S}_j^k, A_j\}_{1\leq j\leq t-1} \mid \mathbb{S}_t^k, A_t, \\
        &\mathbb{S}_{t+1} \ind \{\mathbb{S}_j, A_j\}_{1\leq j\leq t-1} \mid \mathbb{S}_t, A_t.
    \end{align*}
    So far, we have shown that the Markov property holds for $\mathbb{X}_{t}$. In the next step, we show the connection between $\mathbb{S}_{t}$ and $\mathcal{S}_t$. The key assumption we use here is Assumption~\ref{asm:unmeasured_confounders}. We use mathematical induction to show that the Markov property holds for $\mathcal{S}_t$.
    
    \noindent \textbf{First}, assume that for $t\leq l$, we have
    \begin{align}
    	&P(S_{t}^{Z\leftarrow z^{(k)},\bar{A}_{t-1}\leftarrow \bar{a}_{t-1}}(\bar{U}_{t}(H_{t})) \mid S_{1}^{Z\leftarrow z^{(k)}}(U_1(H_1)),\dots, S_{t-1}^{Z\leftarrow z^{(k)},\bar{A}_{t-2}\leftarrow \bar{a}_{t-2}}(\bar{U}_{t-1}(H_{t-1})))& \nonumber \\
    	= &P(S_{t}^{Z\leftarrow z^{(k)}}(\bar{U}_{t}(H_{t})) \mid S_{t-1}^{Z\leftarrow z^{(k)}}(\bar{U}_{t-1}(H_{t-1})), A_{t-1} = a_{t-1}) \nonumber \\
        = &P(\mathbb{S}_t^k \mid \mathbb{S}_{t-1}^k, A_{t-1} = a_{t-1}). \label{eq:tl}
    \end{align}
    
    \noindent It is obvious that \ref{eq:tl} holds when $t=2$ due to Assumption~\ref{asm:unmeasured_confounders},
    \begin{align}
        &P(S_{2}^{Z\leftarrow z^{(k)},A_{1}\leftarrow a_{1}}(\bar{U}_{2}(H_2)) \mid S_{1}^{Z\leftarrow z^{(k)}}(U_1(H_1))) \nonumber \\
        = &P(S_{2}^{Z\leftarrow z^{(k)}}(\bar{U}_{2}(H_2)) \mid S_{1}^{Z\leftarrow z^{(k)}}(U_1(H_1)), A_1 = a_1) \nonumber \\
        = &P(\mathbb{S}_{2}^{k} \mid \mathbb{S}_{1}^{k}, A_1 = a_1).\label{eq:t1}
    \end{align}
    
    \noindent \textbf{Second}, for $t=l+1$, we have
    \begin{align}
    	&P(S_{t}^{Z\leftarrow z^{(k)},\bar{A}_{t-1}\leftarrow \bar{a}_{t-1}}(\bar{U}_{t}(H_{t})),\cdots,S_{1}^{Z\leftarrow z^{(k)},A_{1}\leftarrow a_{1}}(\bar{U}_{2}(H_{2})) \mid S_{1}^{Z\leftarrow z^{(k)}}(\bar{U}_{1}(H_{1}))) \nonumber \\
    	=&P(S_{t}^{Z\leftarrow z^{(k)},A_2\dots A_{t-1}\leftarrow a_2\dots a_{t-1}}(\bar{U}_{t}(H_{t})),\cdots,S_{2}^{Z\leftarrow z^{(k)}}(\bar{U}_{t-1}(H_{t-1})) \mid S_{1}^{Z\leftarrow z^{(k)}}(\bar{U}_{1}(H_{1})),A_1=a_1).\label{eq:t_tm1}
    \end{align}
    Combine \ref{eq:t_tm1} with \ref{eq:t1}, we have
    \begin{align}
    	&P(\{S_{j}^{Z\leftarrow z^{(k)},\bar{A}_{j-1}\leftarrow \bar{a}_{j-1}}(\bar{U}_{j}(H_{j}))\}_{j=3,\dots,t} \mid S_{2}^{Z\leftarrow z^{(k)},A_{1}\leftarrow a_{1}}(\bar{U}_{2}(H_{2})), S_{1}^{Z\leftarrow z^{(k)}}(\bar{U}_{1}(H_{1}))) \nonumber \\
    	=&P(\{S_{j}^{Z\leftarrow z^{(k)},A_2\dots A_{j-1}\leftarrow a_2,\dots a_{j-1}}(\bar{U}_{j}(H_{j}))\}_{j=3,\dots,t} \mid S_{2}^{Z\leftarrow z^{(k)}}(\bar{U}_{2}(H_{2})), S_{1}^{Z\leftarrow z^{(k)}}(\bar{U}_{1}(H_{1})),A_1=a_1).
    \end{align}
    By repeating this procedure up to time $t-1$, we have
    \begin{align}
    	&P(S_{t}^{Z\leftarrow z^{(k)},\bar{A}_{t-1}\leftarrow \bar{a}_{t-1}}(\bar{U}_{t}(H_{t})) \mid \{S_{j}^{Z\leftarrow z^{(k)},\bar{A}_{j-1}\leftarrow \bar{a}_{j-1}}(\bar{U}_{j}(H_{j})\}_{j=2,\dots,t-1}, S_{1}^{Z\leftarrow z^{(k)}}(\bar{U}_{1}(H_{1}))) \nonumber\\
    	=&P(S_{t}^{Z\leftarrow z^{(k)},A_{t-1}\leftarrow a_{t-1}}(\bar{U}_{t}(H_{t})) \mid \{S_{j}^{Z\leftarrow z^{(k)}}(\bar{U}_{j}(H_{j})\}_{j=2,\dots,t-1}, S_{1}^{Z\leftarrow z^{(k)}}(\bar{U}_{1}(H_{1})),\bar{A}_{t-2}=\bar{a}_{t-2}) \nonumber \\
    	=&P(S_{t}^{Z\leftarrow z^{(k)}}(\bar{U}_{t}(H_{t})) \mid \{S_{j}^{Z\leftarrow z^{(k)}}(\bar{U}_{j}(H_{j})\}_{j=2,\dots,t-1}, S_{1}^{Z\leftarrow z^{(k)}}(\bar{U}_{1}(H_{1})),\bar{A}_{t-1}=\bar{a}_{t-1}) \nonumber \\
            =&P(\mathbb{S}_t^k \mid \{\mathbb{S}_{j}^{k}\}_{j=1,\dots,t-1},\bar{A}_{t-1}=\bar{a}_{t-1}) \nonumber \\
    	=&P(\mathbb{S}_t^k \mid \mathbb{S}_{j}^{t-1},A_{t-1}=a_{t-1})
    \end{align}
\noindent The first and second equation are due to Assumption~\ref{asm:unmeasured_confounders}, the third equation is due to the Markov property of $\mathbb{S}_t^k$. Therefore, we show that the Markov property holds for $\mathcal{S}_t^k$. Similarly, we can show that the Markov property holds for $\mathcal{S}_t$. Let $\tilde{\mathbb{R}}_t$ to be the weighted sum of $\mathbb{R}_t^k$,
\begin{align*}
    \tilde{\mathbb{R}}_t = \sum_k p_k \mathbb{R}_t^k,
\end{align*}
where $\{p_k\}_{k=1,\dots,K}$ are fixed weights with $\sum_k p_k = 1$. Note that we have 
\begin{align*}
    \mathbb{S}_t, \tilde{\mathbb{R}}_{t-1} \indep \{\tilde{\mathbb{R}}_{j-1},\mathbb{S}_j, A_j\}_{j=2,\dots,t-2} \mid \mathbb{S}_{t-1}, A_{t-1}.
\end{align*}
By applying Lemma 1 of \citep{shi2020does}, we show that there exists a stationary deterministic policy $\pi(\cdot|\mathbb{S}_t)$ such that it maximize the integrated value function defined as follow,
\begin{align}
    J(\pi) =& \int_{s} E \left[\sum_{t=1}^\infty \gamma^{t-1} \tilde{\mathbb{R}}_t \mid \mathbb{S}_1=s \right] P(\mathbb{S}_1=x) ds \nonumber \\
    =& \int_{s} E \left[\sum_{t=1}^\infty \gamma^{t-1} \sum_k \tilde{\mathbb{R}}_t^k p_k \mid \mathbb{S}_1=s \right] P(\mathbb{S}_1=s) ds \nonumber \\
    =& \int_{s} \sum_k E \left[\sum_{t=1}^\infty \gamma^{t-1}  \tilde{\mathbb{R}}_t^k \mid \mathbb{S}_1=s,Z = z_k \right] p_k P(\mathbb{S}_1=s) ds \nonumber \\
    =& \int_{s} \sum_k V_0^\pi(s,z_k) p_k P(\mathbb{S}_1=s) ds, \label{eq:J}
\end{align}
where $V_0^\pi(\cdot,\cdot)$ is the value function for original CMDP. Note that we have $\mathbb{S}_1 \indep Z$ due to the definition of counterfactual state. If $p_k$ match the distribution of $P(Z=z_k)$ in the target population, maximizing \ref{eq:J} is the same as maximizing
\begin{align*}
    J_0(\pi) = \int_{s,z} V_0^\pi(s,z) P(S_1=s,Z=z) dsdz,
\end{align*}
which is the same as integrated value function of original CMDP with observed reward in the target population.

\end{proof}

\section{Proof of Theorem~\ref{thm:regret}}

\begin{assumption}[Hypothesis Class] \label{asm:hypothesis}
\begin{equation*}
    \mathcal{F} = \{w^T \phi(s,a): w \in \mathbb{R}^d, \|w\| \leq \mathcal{B}\},
\end{equation*}
where $\phi$ is a feature map $\mathcal{S} \times \mathcal{A} \rightarrow \mathbb{R}^d$ with $\|\phi(s,a)\|_\infty \leq 1$. 
\end{assumption}

\begin{assumption}[Completeness] \label{asm:completeness}
    Let $\mathcal{T}$ be the Bellman optimality operator: for $f: \mathcal{S} \times \mathcal{A} \rightarrow \mathbb{R}$,
    \begin{equation*}
        \mathcal{T} f(s,a) = r(s,a) + \gamma \E_{s' \sim P(\cdot \mid s,a)} \max_{a' \in \mathcal{A}} f(s',a').
    \end{equation*}
    Assume for any $f \in \mathcal{F}$, $\mathcal{T} f \in \mathcal{F}$.
\end{assumption}

\begin{assumption}[Feature Coverage] \label{asm:coverage}
There exists a constant $\lambda_0 > 0 $ such that
\begin{align*}
    \lambda_{min} (\E_{s,a \sim \mu_b}[\phi(s,a) \phi(s,a)^T]) \geq \lambda_0.
\end{align*}

\end{assumption}
\begin{assumption}[Lipschitz Continuous Features] \label{asm:lipschitz}
    There exists a constant $L>0$ such that 
    \begin{align*}
        \|\phi_i(s,a) - \phi_i(s',a)\| \leq L\|s-s'\|,
    \end{align*}
    for any $a \in \mathcal{A}$ and $i=1,\dots,d$.
\end{assumption}

\begin{assumption}[Bounded Reward] \label{asm:bounded_reward}
    There exists a constant $R_{\max} > 0$ such that $|r| \leq R_{\max}$.
\end{assumption}

\begin{proof}
Let $w_f$ denote the parameter estimate when the value function is $f$, namely,
\begin{align} \label{eq:omega_f}
    w_f = \left[\sum_{i=1}^n \phi(s_i,a_i) \phi (s_i,a_i)^\top\right]^{-1} \left [ \sum_{i=1}^n \phi(s_i,a_i) \left (r_i + \gamma \max_{a^\prime \in \mathcal{A}} f(s_i^\prime,a^\prime) \right)\right ].
\end{align}
Let $\hat{w}_f$ be the corresponding parameter estimate when the $s_i$, $s_i^\prime$ and $r_i$ are replaced by estimated $\hat{s}_i$, $\hat{s}_i^\prime$ and $\hat{r}_i$, namely,
\begin{align} \label{eq:omega_f_hat}
    \hat{w}_f = \left[\sum_{i=1}^n \phi(\hat{s}_i,a_i) \phi (\hat{s}_i,a_i)^\top\right]^{-1} \left [ \sum_{i=1}^n \phi(\hat{s}_i,a_i) \left (\hat{r}_i + \gamma \max_{a^\prime \in \mathcal{A}} f(\hat{s}_i^\prime,a^\prime) \right)\right ].
\end{align}
According to Theorem 8 in \citet{hu2024fast}, we know that
\begin{align}
	\|Q^* - \hat{f}_B\|_\infty  \leq \sum_{t=0}^{B-1} \gamma^t \|\hat{f}_{B-t} - \mathcal{T} \hat{f}_{B-t-1}\|_\infty + \frac{\gamma^B R_{max}}{1-\gamma}. \label{eq:q_bound}
\end{align}
The second term is a diminishing term as the number of iterations $B$ increase. Therefore, we aim to characterize the first term, which is a weighted sum of one-step estimation errors. For any $f \in \mathcal{F}$, one step estimation error can be bounded by,
\begin{align}
    \|\hat{w}_f^\top \phi - \mathcal{T}f\|_\infty &\leq \|\hat{w}_f^\top \phi - w_f^\top \phi + w_f^\top \phi - \mathcal{T}f\|_\infty \nonumber \\
    &\leq \|\hat{w}_f^\top \phi - w_f^\top \phi \|_\infty + \|w_f^\top \phi - \mathcal{T}f\|_\infty \nonumber \\
    &\leq \|\hat{w}_f^\top - w_f^\top\|_\infty \cdot \|\phi\|_\infty + \|w_f^\top \phi - \mathcal{T}f\|_\infty.
\end{align}
The second term can be bounded by Lemma 7 in \citet{hu2024fast}. The first term characterizes the estimation error when the state and reward are estimated. To simplify the notation, we denote

\begin{align}
    A &= \sum_{i=1}^n \phi(s_i,a_i) \phi (s_i,a_i)^\top, \nonumber \\
    \mathbf{b} &= \sum_{i=1}^n \phi(s_i,a_i) \left (r_i + \gamma \max_{a^\prime \in \mathcal{A}} f(s_i^\prime,a^\prime) \right), \nonumber \\
    C &= \sum_{i=1}^n \phi(\hat{s}_i,a_i) \phi (\hat{s}_i,a_i)^\top, \nonumber \\
    \mathbf{d} &= \sum_{i=1}^n \phi(\hat{s}_i,a_i) \left (\hat{r}_i + \gamma \max_{a^\prime \in \mathcal{A}} f(\hat{s}_i^\prime,a^\prime) \right). \nonumber
\end{align}
And we can bound the estimation error,
\begin{align}
    \left\|\hat{w}_f - w_f\right\|_\infty &= \left\| A^{-1}\mathbf{b} - C^{-1}\mathbf{d} \right\|_\infty \nonumber\\
    &= \left\| A^{-1}\mathbf{b} - A^{-1}\mathbf{d} + A^{-1}\mathbf{d} - C^{-1}\mathbf{d} \right\|_\infty \nonumber\\
    &\leq \left\|A^{-1}\right\|_2 \left\|\mathbf{b}-\mathbf{d}\right\|_\infty +  \left\|A^{-1} - C^{-1}\right\|_2 \left\|\mathbf{d}\right\|_\infty \nonumber\\
    &\leq \left\|A^{-1}\right\|_2 \left\|\mathbf{b}-\mathbf{d}\right\|_\infty +  \left\|A^{-1} \right\|_2 \left\|A-C\right\|_2 \left\| C^{-1}\right\|_2 \left\|\mathbf{d}\right\|_\infty \label{eq:abcd}. 
\end{align}
Next step, we bound each term in (\ref{eq:abcd}). 

\paragraph{Bound $\|A^{-1}\|_2$:} By Lemma 7 in \citet{hu2024fast}, we have 
\begin{align*}
    P(\lambda_{\min}(A) \geq n\lambda_0/2) \leq d \exp(-n\lambda_0/8).
\end{align*}
When the event $\lambda_{\min}(A) \geq n\lambda_0/2$ holds, we have 
\begin{align}
	\|A^{-1}\|_2 \leq \frac{2}{n\lambda_0}.
\end{align}

\paragraph{Bound $\|A-C\|_2$:} To bound this term, we first consider bounding element-wise difference between $A$ and $C$,
\begin{align}
    \left|A_{jk}-C_{jk}\right| &= \left|\sum_{i=1}^n \phi_j(s_i,a_i)\phi_k(s_i,a_i) - \phi_j(\hat{s}_i,a_i)\phi_k(\hat{s}_i,a_i)\right| \nonumber \\ 
    &= \left|\sum_{i=1}^n \phi_j(s_i,a_i)(\phi_k(s_i,a_i) - \phi_k(\hat{s}_i,a_i)) + (\phi_j(s_i,a_i) - \phi_j(\hat{s}_i,a_i))\phi_k(\hat{s}_i,a_i)\right| \nonumber \\
    &\leq \left|\sum_{i=1}^n (\phi_k(s_i,a_i) - \phi_k(\hat{s}_i,a_i)) + (\phi_j(s_i,a_i) - \phi_j(\hat{s}_i,a_i))\right| \nonumber \\
    &\leq 2nL\epsilon \nonumber.
\end{align}
Therefore, we have
\begin{align}
	\|A-C\|_2 &\leq \|A-C\|_F = \sqrt{\sum_{j=1}^d \sum_{k=1}^{d} \left|A_{j,k} - C_{j,k}\right|^2} \leq 2ndL\epsilon
\end{align}
\paragraph{Bound $\|C^{-1}\|_2$:}
For any semi positive-definite matrix $A$ and $C$, we have
\begin{align*}
	|\lambda_{\min}(A) - \lambda_{\min}(C)| \leq \|A-C\|_2 \leq 2ndL\epsilon.
\end{align*}
When the event $\lambda_{\min}(A) \geq n\lambda_0/2$ holds, we have 
\begin{align*}
	\lambda_{\min}(C) \geq n\lambda_0/2 - 2ndL\epsilon,
\end{align*}
and we have
\begin{align}
	\|C^{-1}\|_2 \leq \frac{2}{n\lambda_0 - 4ndL\epsilon}.
\end{align}

\paragraph{Bound $\|\mathbf{b} - \mathbf{d}\|_\infty$:} We can bound $\|\mathbf{b} - \mathbf{d}\|_\infty$ by

\begin{align}
    \|\mathbf{b} - \mathbf{d}\|_\infty &= \bigg\|\sum_{i=1}^n\phi(\hat{s}_i,a_i)(r_i + \gamma \max_{a'\in\mathcal{A}} f(s_i^\prime,a_i^\prime) - \hat{r}_i - \gamma \max_{a'\in \mathcal{A}} f(\hat{s}_i^\prime,a_i^\prime)) \nonumber \\
    &+ \sum_{i=1}^n (\phi(s_i,a_i) - \phi(\hat{s}_i,a_i))(r_i + \gamma \max_{a' \in \mathcal{A}} f(s_i^\prime,a_i))\bigg\|_\infty \nonumber \\
    &\leq \bigg\|\sum_{i=1}^n\phi(\hat{s}_i,a_i)(r_i + \gamma \max_{a'\in\mathcal{A}} f(s_i^\prime,a_i^\prime) - \hat{r}_i - \gamma \max_{a'\in \mathcal{A}} f(\hat{s}_i^\prime,a_i^\prime))\bigg\|_\infty \nonumber \\
    &+ \bigg\|\sum_{i=1}^n (\phi(s_i,a_i) - \phi(\hat{s}_i,a_i))(r_i + \gamma \max_{a' \in \mathcal{A}} f(s_i^\prime,a_i))\bigg\|_\infty \nonumber \\
    & \leq \sum_{i=1}^n|r_i - \hat{r}_i|\bigg\|\phi(\hat{s}_i,a_i)\bigg\|_\infty + \gamma \sum_{i=1}^n \bigg|\max_{a'\in\mathcal{A}} f(s_i^\prime,a_i^\prime)- \max_{a'\in \mathcal{A}} f(\hat{s}_i^\prime,a_i^\prime)\bigg|\bigg\|\phi(\hat{s}_i,a_i) \bigg\|_\infty \nonumber \\
    &+ \sum_{i=1}^n \bigg|(r_i + \gamma \max_{a' \in \mathcal{A}} f(s_i^\prime,a_i))\bigg| \bigg\|(\phi(s_i,a_i) - \phi(\hat{s}_i,a_i))\bigg\|_\infty \nonumber \\
    & \leq \sum_{i=1}^n|r_i - \hat{r}_i|\bigg\|\phi(\hat{s}_i,a_i)\bigg\|_\infty + \gamma \sum_{i=1}^n L \|s_i - \hat{s}_i\|_\infty \bigg\|\phi(\hat{s}_i,a_i) \bigg\|_\infty \nonumber \\
    &+ \sum_{i=1}^n \bigg|(r_i + \gamma \max_{a' \in \mathcal{A}} f(s_i^\prime,a_i))\bigg| \bigg\|\phi(s_i,a_i) - \phi(\hat{s}_i,a_i)\bigg\|_\infty \nonumber \\
    &\leq (1 + \gamma L + \frac{R_{\max}L}{1-\gamma}) n\epsilon.
\end{align}
\paragraph{Bound $\|d\|_\infty$:} To bound $\|d\|_\infty$, we consider 
\begin{align}
    \|\mathbf{d}\|_\infty &\leq \|\mathbf{b}-\mathbf{d}\|_\infty + \|\mathbf{b}\|_\infty \nonumber \\
    &\leq (1 + \gamma L + \frac{R_{\max}L}{1-\gamma}) n\epsilon + \|\sum_{i=1}^n \phi(s_i,a_i) \left (r_i + \gamma \max_{a^\prime \in \mathcal{A}} f(s_i^\prime,a^\prime) \right)\|_\infty \nonumber \\
    &\leq (1 + \gamma L + \frac{R_{\max}L}{1-\gamma}) n\epsilon + \sum_{i=1}^n |\left (r_i + \gamma \max_{a^\prime \in \mathcal{A}} f(s_i^\prime,a^\prime) \right)|\| \phi(s_i,a_i) \|_\infty \nonumber  \\
    &\leq (1 + \gamma L + \frac{R_{\max}L}{1-\gamma}) n\epsilon + \frac{n R_{\max}}{1-\gamma}.
\end{align}

\paragraph{Bound $\left\|\hat{w}_f - w_f\right\|_\infty$:} With bound for each term, we have, 
\begin{align}
    \left\|\hat{w}_f - w_f\right\|_\infty &\leq \left\|A^{-1}\right\|_2 \left\|\mathbf{b}-\mathbf{d}\right\|_\infty +  \left\|A^{-1} \right\|_2 \left\|A-C\right\|_2 \left\| C^{-1}\right\|_2 \left\|\mathbf{d}\right\|_\infty \nonumber \\
    &\leq \frac{2}{n\lambda_0} (1 + \gamma L + \frac{R_{\max}L}{1-\gamma}) n\epsilon \nonumber\\
    &+ \frac{2}{n\lambda_0} * 2ndL\epsilon * \frac{2}{n\lambda_0 - 4ndL\epsilon} * ((1 + \gamma L + \frac{R_{\max}L}{1-\gamma}) n\epsilon + \frac{n R_{\max}}{1-\gamma}) \nonumber\\
    & = \frac{2\epsilon}{\lambda_0 - 4dL\epsilon} \left(1 + \gamma L + \frac{R_{\max}L}{1-\gamma}\right) + \frac{8dL\epsilon R_{\max}}{\lambda_0 (\lambda_0 - 4dL\epsilon)(1-\gamma)} \nonumber \\
    & \leq C_1 \frac{dLR_{\max}\epsilon}{\lambda_0(\lambda_0-4dL\epsilon)(1-\gamma)}, \label{eq:bound_what_w}
\end{align}
where we used the fact that $\lambda_0 \leq 1$ and $C_1$ is a positive constant.

\noindent Combined \ref{eq:bound_what_w} with Lemma 7 in \citet{hu2024fast}, we have
\begin{align}
    &P\left(\sup_f \|w_f^T \phi - \mathcal{T}f\|_\infty \geq \delta\right) \leq 6d\exp\left(-\frac{\lambda_0^2}{5184d^2(R_{\max}+\mathcal{B})^2}n\delta^2\right), \label{eq:p_fqi} \\
    &P\left(\sup_f\|\hat{w}_f^T \phi - w_f^T \phi\|_\infty \geq \frac{C_1dLR_{\max}\epsilon}{\lambda_0(\lambda_0-4dL\epsilon)(1-\gamma)}\right) \leq d \exp(-n\lambda_0/8). \label{eq:p_estimation}
\end{align}
Here we assume that there exists a positive constant $C^*$ such that $\mathcal{B} < \frac{C^*R_{\max}}{1-\gamma}$ since the optimal Q function is on the order of $\frac{R_{\max}}{1-\gamma}$ and $\|\phi\|_\infty \leq 1$. We also use the fact that $d<n$. Assume $\kappa$ is a sufficiently large positive constant, thus we can simplify \ref{eq:p_fqi}:
\begin{align}
    P\left(\sup_f \|w_f^T \phi - \mathcal{T}f\|_\infty \geq \frac{C_2dR_{\max}\kappa\log(n)}{(1-\gamma)\lambda_0\sqrt{n}}\right) \leq n^{-\kappa}, \label{eq:p_fqi_2}
\end{align}
where $C_2$ is a positive constant. Applying Bonferroni inequality to \ref{eq:p_estimation} and \ref{eq:p_fqi_2}, we have
\begin{align}
    &P\left(\sup_f \|\hat{w}_f^T \phi - \mathcal{T}f\|_\infty \leq  \frac{C_1dLR_{\max}\epsilon}{\lambda_0(\lambda_0-4dL\epsilon)(1-\gamma)} +  \frac{C_2dR_{\max}\kappa\log(n)}{(1-\gamma)\lambda_0\sqrt{n}} \right) \leq 1 - n^{-\kappa} - d\exp(-n\lambda_0/8).
\end{align}
According to \ref{eq:q_bound} and Lemma 13 in \citet{chen2019information}, we have the bound for regret
\begin{align}
    P\left(\nu^* - \nu^{\hat{\pi}_{\hat{f}_B}} \leq\frac{ C_1 dLR_{\max}\epsilon}{\lambda_0(\lambda_0-4dL\epsilon)(1-\gamma)^3} + \frac{ C_2dR_{\max}\kappa\log(n)}{\lambda_0\sqrt{n}(1-\gamma)^3} + \frac{\gamma^B R_{\max}}{(1-\gamma)^2}\right)& \nonumber \\
    \leq 1 - n^{-\kappa} - d\exp(-n\lambda_0/8)&,
\end{align}
where $\nu^\pi = \E_s^\pi[\sum_{h=1}^\infty \gamma^{h-1} r_h]$.
\end{proof}

\section{Proof of Theorem~\ref{thm:cf_metric}}
\begin{assumption}[Margin] \label{asm:margin} 
    Assume there exist some constants $\alpha$ such that 
    \begin{align*} 
        \mathbb{P}\left\{s \in S: \max_a Q^{opt}(s,a) - \max_{a'\in A - \arg\max_a Q^{opt}(s,a)} Q^{opt}(s,a') \leq \epsilon\right\} = O\left(\epsilon^\alpha\right).
    \end{align*}
    where $\mathbb{P}$ is the initial state  distribution.
    
\end{assumption}
\begin{proof}
For each $\tilde{s}$, we want to derive the bound for $\|\hat{\pi}(\hat{\tilde{s}}_{z'}) - \hat{\pi}(\hat{\tilde{s}}_{z''})\|_2$, where $\hat{\tilde{s}}_{z}$ is the estimated $\tilde{s}$ when $Z=z$ is observed. Note that by the definition of counterfactuals, the oracle value for $\hat{\tilde{s}}_{z'}$ and $\hat{\tilde{s}}_{z''}$ is $\tilde{s}$. To simplify the notation, we will omit $\tilde{\boldsymbol{\cdot}}$ in the proof below.

\noindent By triangle inequality, we have
\begin{align}
    \|\hat{\pi}(\hat{s}_{z'}) - \hat{\pi}(\hat{s}_{z''})\|_2 \leq& \|\hat{\pi}(\hat{s}_{z'}) - \hat{\pi}(s) \|_2 + \|\hat{\pi}(s)  - \hat{\pi}(\hat{s}_{z''})\|_2. \label{eq:pihat_zp_zpp}
\end{align}
The first and second terms in \ref{eq:pihat_zp_zpp} are similar. Therefore, we only consider the first term in the following proof and another term can be treated similarly. By triangle inequality,
\begin{align}
        & \E\|\hat{\pi}(\hat{s}_{z}) - \hat{\pi}(s) \|_2 \leq& \E\|\hat{\pi}(s) - \pi^*(s) \|_2 + \E\|\hat{\pi}(\hat{s}_{z}) - \pi^*(\hat{s}_{z}) \| + \E\|\pi^*(\hat{s}_{z}) - \pi^*(s) \|_2.
\end{align}
Here we assume that $\hat{s}_{z}$ and $s$ has the same support $S$.
\paragraph{Bound $\E_{s}\|\hat{\pi}(s) - \pi^*(s) \|_2$:} Using Theorem~\ref{thm:regret}, we have $P(\mathcal{A}_0) \geq 1 - n^{-\kappa} - d\exp(-n\lambda_0/8)$, where
\begin{align*}
    \mathcal{A}_0 = \{\|Q^* - \hat{f}\|_\infty \leq \xi\},
\end{align*} 
where 
\begin{align}
    \xi = \frac{ C_1 dLR_{\max}\epsilon}{\lambda_0(\lambda_0-4dL\epsilon)(1-\gamma)^2} + \frac{ C_2dR_{\max}\kappa\log(n)}{\lambda_0\sqrt{n}(1-\gamma)^2} + \frac{\gamma^B R_{\max}}{1-\gamma}. \label{eq:fqi_error_bound}
\end{align}
For any $s \in S$, suppose
	\begin{align} \label{eq:thm3_x}
		\max_a Q^{opt}(s,a) - \max_{a'\in A - \arg\max_a Q^{opt}(s,a)} Q^{opt}(s,a') > 2 \xi.
	\end{align}
Under the event defined in $\mathcal{A}_0$, we have
	\begin{align*}
		\max_a \hat{f} (s,a) - \max_{a'\in A - \arg\max_a Q^{opt}(s,a)} \hat{f}(s,a') > 0,
	\end{align*}
and hence
	\begin{align*}
		\{a \in \mathcal{A}: \hat{\pi}(a|s) = 1\} \subseteq \arg \max_{a\in \mathcal{A}} Q^{opt}(s,a).
	\end{align*}
Thus, we have
	\begin{align*}
		\|\hat{\pi}(s) - \pi^*(s) \|_2 = 0.
	\end{align*}
Let $\mathbb{S}$ denote the set of $s$ that satisfies \eqref{eq:thm3_x}. It follows that
	\begin{align}
		&\int_s \|\hat{\pi}(s) - \pi^*(s) \|_2  \mathbb{I}(\mathcal{A}_0) \mathbb{P}(ds) \nonumber \\
		= &\int_x \|\hat{\pi}(s) - \pi^*(s) \|_2  \mathbb{I}(\mathcal{A}_0)\mathbb{I}(s \in \mathbb{S}^c)\mathbb{P}(ds) \nonumber \\ 
		\leq & \int_s \mathbb{I}(s \in \mathbb{S}^c)\mathbb{P}(ds) \nonumber \\
		= & (2\xi)^\alpha
	\end{align}

\paragraph{Bound $\E\|\hat{\pi}(\hat{s}_z) - \pi^*(\hat{s}_z) \|_2$:} Using similar idea for $\E_{s}\|\hat{\pi}(s) - \pi^*(s) \|_2$, we have
\begin{align}
    \E \|\hat{\pi}(\hat{s}_z) - \pi^*(\hat{s}_z) \|_2 \leq (2\xi)^\alpha
\end{align}

\paragraph{Bound $\E\|\pi^*(\hat{s}_z) - \pi^*(s) \|_2$:} By definition of $\epsilon$, we have $\max_s \|\hat{s}_z - s\| \leq \epsilon$. For any $s \in S$, suppose
\begin{align} \label{eq:thm3_xepsilon}
	\max_a Q^{opt}(s,a) - \max_{a'\in A - \arg\max_a Q^{opt}(s,a)} Q^{opt}(s,a') > 2 L \epsilon.
\end{align}
We have
\begin{align}
    \max_a Q^{opt}(\hat{s}_z,a) - \max_{a'\in A - \arg\max_a Q^{opt}(s,a)} Q^{opt}(\hat{s}_z,a') > 0.
\end{align}
Thus we have
    \begin{align*}
	\|\pi^*(\hat{s}_{z}) - \pi^*(s) \|_2 = 0.
    \end{align*}
Let $\mathbb{S}$ denote the set of $s$ that satisfies \eqref{eq:thm3_xepsilon}. It follows that
	\begin{align}
		&\E_{\hat{s}_{z}}\left [\int_{s} \|\pi^*(\hat{s}_{z}) - \pi^*(s) \|_2   \mathbb{P}(ds) \right] \nonumber \\
		= &\E_{\hat{s}_{z}}\left [\int_x \|\pi^*(\hat{s}_{z}) - \pi^*(s) \|_2 \mathbb{I}(s \in \mathbb{S}^c)\mathbb{P}(ds) \right ] \nonumber \\ 
		\leq & \E_{\hat{s}_{z}}\left [\int_s \mathbb{I}(s \in \mathbb{S}^c)\mathbb{P}(ds) \right ] \nonumber \\
		= & (2L\epsilon)^\alpha.
	\end{align}
Combined all the results together, we have
\begin{align}
    P\left\{\E\|\hat{\pi}(\hat{s}_{z'}) - \hat{\pi}(\hat{s}_{z''}) \|_2 \leq 2^{\alpha+1} \xi^\alpha +2^\alpha (L\epsilon)^\alpha \right\} \geq 1 - n^{-\kappa} - d\exp(-n\lambda_0 / 8). 
\end{align}
\end{proof}

\section{Details on numerical experiments} \label{apdx:num_exp}

\subsection{Implementation details}
To implement the Algorithm~\ref{alg:cf:1}, we use a neural network with $[64,64]$ hidden layer to approximate the transition kernels. The transition kernels are fitted separately for each value of the sensitive attribute to account for potential heterogeneity. Alternative machine learning models can be used to fit the transition kernels depending on the specific characteristics of their data and application domain. The models are fitted using a learning rate of $0.005$, a batch size of $512$, and a maximum of $1000$ epochs. We use the mean squared error as the loss function. To prevent overfitting, the dataset was split into training and testing sets with an $80/20$ ratio.  Early stopping was implemented to terminate training if the test loss did not improve by more than $0.01$ for a consecutive period of $10$ epochs. We adopt fitted Q iteration (FQI) algorithm for offline policy learning. The Q-functions are modeled by neural network with a single hidden layer $[32]$. The FQI algorithm is trained for $100$ iterations, and within each iteration, the network was optimized for $500$ gradient descent steps using a learning rate of $0.1$. The Adam optimizer \citep{kingmaBa2015} is used throughout the experiments. The experiments are conducted on an internal cluster with 60-core Intel Xeon Gold $6230$ CPU clocked at $2.10$ GHz and 192 GB memory.  A discount factor of $\gamma=0.9$ is used across all experiments.

\subsection{Synthetic data}\label{apdx:syn}

\subsubsection{Data generating mechanism for linear setting}
\begin{align*}
    S_1 &= -0.3 + 1.0 \delta Z + U_{S_1}, \\
    S_t &= -0.3 + 1.0 \delta (Z-0.5) + 0.5 S_{t-1} + 0.4 (A_{t-1} - 0.5) + 0.3 S_{t-1}(A_{t-1} - 0.5) \\
        &+ 0.3 \delta S_{t-1}(Z-0.5) +0.4 \delta (Z-0.5)(A_{t-1} - 0.5) + U_{S_t},\\
    R_t &= -0.3 + 0.3 S_t + 0.5 \delta Z + 0.5 A_t + 0.2 \delta S_tZ + 0.7 S_t A_t - 1.0 \delta ZA_t,
\end{align*}
where $U_{S_t} \sim \mathcal{N}(0,1)$, $Z$ takes a binary value in $\{0,1\}$ with equal probability. Behavior policy is set as $\mathbb{P}(A_t = 1 \mid Z ) = \text{expit}(-1.39 + 2.77Z)$. $\delta$ is used to control the impact of sensitive attribute on state variables. When $\delta=0$, the state variables $\{S_t\}_{t\geq1}$ are independent of sensitive attribute $Z$; the effect of $Z$ on $\{S_t\}_{t\geq1}$ increases as $\delta$ increases.

\subsubsection{Data generating mechanism for non-linear setting}
\begin{align*}
    S_1 &= -0.7 + 0.8 Z + U_{S_1}, \\
    S_t &= -1.0 + 0.8 \delta Z + 0.25 (\sin{S_{t-1}}+\cos{S_{t-1}}) + 0.4 (A_{t-1} - 0.5) \\
    &+ 0.15 (\sin{S_{t-1}}+\cos{S_{t-1}})(A_{t-1} - 0.5) \\
    &+ 0.15 \delta (\sin{S_{t-1}}+\cos{S_{t-1}})Z + 0.4 \delta Z(A_{t-1} - 0.5) + U_{S_t},\\
    R_t &= -0.2 + 0.3 S_t + 0.8 \delta Z + 0.8 A_t - 0.6 \delta S_tZ - 0.7 S_t A_t - 1.6 \delta ZA_t
\end{align*}
where $U_{S_t} \sim \mathcal{N}(0,1)$, $Z$ takes a binary value in $\{0,1\}$ with equal probability. Behavior policy is set as $P(A_t = 1 \mid Z) = \text{expit}(-1.39 + 2.77Z)$.

\subsection{Semi-synthetic data} \label{apdx:semi}

To mimic the characteristics of real-world data, we first use the data from PowerED study to learn a generative model. The set of state variables $S_t$ consists of weekly pain score and interference at week $t$. The reward $R_t$ is defined as the weekly \texttt{$7$ - weekly self-reported opioid medication risk score}. The action $A_t$ is a ternary variable, including a brief motivational interactive voice response (IVR) call (less than 5 minutes), a longer IVR call (5 to 10 minutes), or a live call with counselor (20 minutes). The sensitive attribute is univariate: gender, age, ethnicity, or education. We binarize these sensitive attributes for convenience.  In this paper, we choose to focus on the sensitive attributes one at a time instead of multivariate, because the latter, although theoretically possible, makes counterfactual state estimation more challenging given limited sample size. In addition, the univariate approach is standard in the current literature as it isolates the role of each sensitive attribute. To account for potential heterogeneity of transition functions among different sensitive attribute groups, for each sensitive attribute group $z$, we fit a separate model $P(S_{t+1},R_t|S_t,A_t,Z=z)$, which follows a multivariate normal distribution with mean $\mu_z(S_t,A_t)$ and diagonal covariance matrix $\Sigma_z$. We use a neural network with $[32, 32]$ structure to approximate each $\mu_z(S_t, A_t)$ for $z\in\mathcal{Z}$; the diagonal elements of $\Sigma_z$ are estimated using the residual variance of each component.


\section{More on real data analysis} \label{apdx:real_data}
In order to calculate the CF metric for each approach, we use the generative model described in Section~\ref{sec:semi} to generate the states and rewards in the counterfactual world. Please refer to Appendix~\ref{apdx:semi} for more details on the generative model. Then we apply each method on the factual and counterfactual worlds separately, and calculate the CF metric defined in Section~\ref{sec:num_exp}. We use fitted Q evaluation (FQE) to evaluate the value for each policy. A neural network with hidden layer $[32]$ is used to model the $Q$ function. The FQE algorithm is trained for $100$ iterations, and within each iteration, the network was optimized for $500$ gradient descent steps using a learning rate of $0.1$. The Adam optimizer is used for optimization.

\begin{figure}
    \centering
    \includegraphics[width=\textwidth]{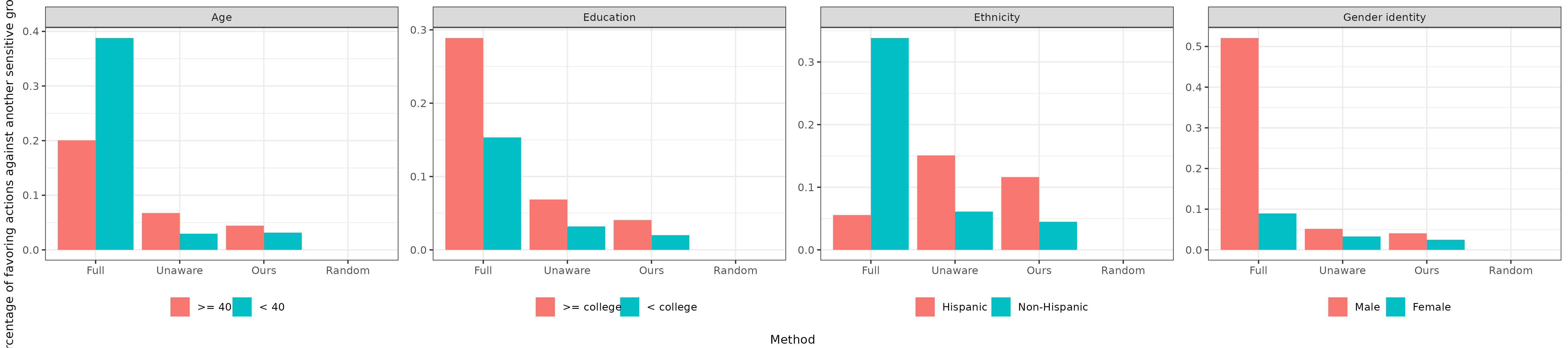}
    \caption{Percentage of favoring actions against another sensitive groups for age, education, ethnicity and gender identity.}
    \label{fig:real_action_percentage}
\end{figure}

\bibliographystyle{agsm}

\bibliography{CFRL_supp}
\makeatletter

\makeatother